\documentclass[lettersize,journal]{IEEEtran}
\usepackage{amsmath,amsfonts}
\usepackage{algorithm}
\usepackage{array}
\usepackage[caption=false,font=footnotesize,labelfont=rm,textfont=rm]{subfig}
\usepackage{textcomp}
\usepackage{makecell}
\usepackage{stfloats}
\usepackage{url}
\usepackage{verbatim}
\usepackage{graphicx}
\usepackage{cite}
\usepackage{colortbl} 
\usepackage{hyperref} 
\usepackage{bigstrut}
\usepackage{caption}
\usepackage{amsmath}
\usepackage{multirow}                 %
\usepackage{multicol}                 %
\usepackage{multirow} 
\usepackage{siunitx}
\usepackage{pifont}       %
\usepackage{flushend}
\usepackage{xcolor}  %
\usepackage{colortbl}  %
\usepackage{algpseudocode}
\usepackage{amssymb}

\usepackage{booktabs,makecell, multirow, tabularx}
\hypersetup{
    colorlinks=true, %
    linkcolor=blue, %
    citecolor=green, %
    urlcolor=blue %
}
\begin{document}

\title{

VL-UniTrack: A Unified Framework with Visual-Language Prompts for UAV-Ground Visual Tracking
}

\author{Boyue Xu, \IEEEmembership{Student Member, IEEE}, Ruichao Hou, \IEEEmembership{Member, IEEE},  Tongwei Ren, \IEEEmembership{Member, IEEE} and Gangshan Wu, \IEEEmembership{Member, IEEE} 

\thanks{This work was supported by the National Natural Science Foundation of China (92582103), the Fundamental and Interdisciplinary Disciplines Breakthrough Plan of the Ministry of Education of China (No. JYB2025XDXM118), the ``111 Center'' (No. B26023), the Research and Innovation Plan for Graduate Students in Jiangsu Province (No. 26CXJH0360), and the Collaborative Innovation Center of Novel Software Technology and Industrialization.
\emph{(Corresponding author: Tongwei Ren.)} }
\thanks{Boyue Xu and Gangshan Wu are with the State Key Laboratory for Novel Software Technology, Nanjing University, Nanjing 210008, China (e-mail: xuby@smail.nju.edu.cn; gswu@nju.edu.cn).}
\thanks{Ruichao Hou is with the School of Artificial Intelligence, China Pharmaceutical University, Nanjing 211198, China (e-mail: rchou@cpu.edu.cn).}
\thanks{Tongwei Ren is with the State Key Laboratory for Novel Software Technology, Nanjing University, Nanjing 210008, China, and also with the School of Artificial Intelligence, China Pharmaceutical University, Nanjing 211198, China (e-mail: rentw@nju.edu.cn).}

\thanks{\textcopyright~2026 IEEE. Author accepted manuscript. Accepted for publication in \emph{IEEE Transactions on Multimedia}.}
}

\markboth{Author Accepted Manuscript, IEEE Transactions on Multimedia}%
{Xu \MakeLowercase{\textit{et al.}}: VL-UniTrack}

\maketitle

\begin{abstract}
UAV-ground visual tracking (UGVT) aims to simultaneously track the same target from both UAV and ground views. However, existing two-stream methods suffer from isolated feature extraction and rely heavily on implicit appearance matching, which often fails to establish reliable cross-view correspondence under drastic viewpoint discrepancies, thereby leading to unstable tracking. To address these issues, we propose VL-UniTrack, a unified framework enhanced by visual-language prompts. By encoding features from both views with a single shared encoder, our method reduces feature isolation between the two views and enables sufficient cross-view interaction. To alleviate the ambiguity caused by appearance-only matching, we further design a visual-language viewpoint prompting module that fuses language descriptions with visual features to generate learnable prompts. These prompts are then injected into a prompt-guided cross-view adapter to facilitate cross-view interaction and guide the learning of view-specific feature representations. In addition, we introduce a confidence-modulated mutual distillation loss to regularize training by suppressing noise propagation. Extensive experiments demonstrate that VL-UniTrack achieves state-of-the-art performance on the latest benchmark. The code is available at \url{https://github.com/xuboyue1999/VL-UniTrack.git}.

\end{abstract}


\section{Introduction}

\IEEEPARstart{U}{AV-ground} visual tracking (UGVT)~\cite{MVCL} aims to simultaneously track the same target from both UAV and ground views. By exploiting the complementary information from these two views, UGVT can alleviate the inherent limitations of single-view tracking, such as occlusion and restricted field of view, and therefore holds great potential for applications such as security surveillance and rescue missions~\cite{UAVUGV1,UAVUGV2}. However, the substantial appearance discrepancy between UAV and ground views makes reliable cross-view feature interaction difficult to establish, rendering traditional tracking methods that rely primarily on visual similarity less effective.
\begin{figure}[!t]
\centering
  \includegraphics[width=0.48\textwidth]{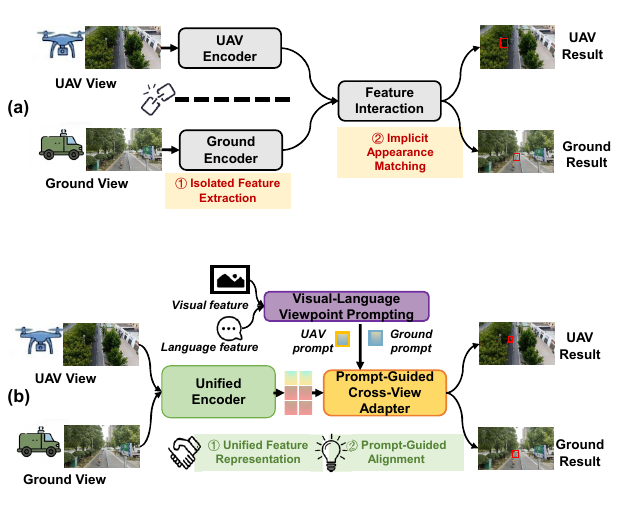}   
  \caption{Framework comparison between existing trackers and the proposed method (VL-UniTrack). (a) Existing tracker: Utilizes a split two-stream architecture with independent encoders, which relies on isolated feature extraction and appearance matching, leading to unreliable feature interaction under drastic viewpoint discrepancies. (b) VL-UniTrack: Uses a unified joint-encoding framework with a single shared encoder, incorporating visual-language viewpoint prompting for explicit view prompts and a prompt-guided cross-view adapter to guide viewpoint-aware interaction within the unified space.}
  \label{fig:intro}  
\end{figure}

To track the same target across different views, existing methods typically adopt a two-stream architecture, where independent visual encoders are used to extract features from the UAV and ground views separately, and feature interaction is performed only at a later stage, as illustrated in Fig.~\ref{fig:intro}(a). Such methods are constrained by two limitations: (1) isolated feature extraction, where separate encoders prevent sufficient interaction between cross-view features; and (2) implicit appearance matching, where feature interaction relies solely on visual similarity and therefore becomes unreliable under drastic cross-view discrepancies, often failing to establish reliable correspondence. The UGVT task presents distinct requirements compared with those of one-stream single-view tracking and aligned multi-modal tracking (\textit{e.g.}, visible-thermal tracking). In UGVT, the UAV and ground views are inherently unaligned and exhibit large viewpoint and appearance discrepancies, so methods developed for these settings may not fully address these cross-view characteristics when directly transferred.

To address these limitations, we propose VL-UniTrack, a unified framework for joint UAV-ground representation learning, as illustrated in Fig.~\ref{fig:intro}(b). First, to address isolated feature extraction, we design a unified joint encoder. Unlike conventional two-stream methods, our framework takes UAV and ground views as joint inputs to a single visual encoder. This design reduces the separation between the two views, allowing cross-view features to be learned within a shared representation space and thereby enabling more thorough interaction. Second, to alleviate the unreliability of appearance-based matching, we incorporate visual-language prompts into the framework. Specifically, we encode view-specific cues as language prompts and fuse them with target appearance features to guide the learning of view-specific feature representations, thereby improving feature discrimination across views and bridging the large viewpoint gap. The unified encoder first establishes a shared cross-view representation through sufficient interaction between the two views, after which viewpoint-specific adaptation uses the corresponding prompts to emphasize the cues required by each tracking head.

Specifically, our method consists of three key modules. First, we introduce a visual-language viewpoint prompting (VLVP) module, which leverages CLIP~\cite{clip} to encode view descriptions and fuse them with visual features, thereby generating learnable prompt tokens for the unified representation. Second, we design a prompt-guided cross-view adapter (PCVA), which injects these prompts into cross-view interaction to explicitly guide the learning of view-specific feature representations. Finally, to regularize the learning of the unified representation, we employ a confidence-modulated mutual distillation (CMD) loss, which dynamically prioritizes the more reliable view and suppresses noise propagation during training. Extensive experiments demonstrate that our method achieves state-of-the-art performance on the latest benchmark.

In summary, our contributions are as follows:
\begin{itemize}
    \item We propose VL-UniTrack, a unified joint-encoding framework for unaligned UAV-ground visual tracking. By employing a single shared encoder for joint feature extraction, our method reduces the feature isolation caused by separate encoding and enables sufficient cross-view interaction.
    
    \item We introduce a visual-language viewpoint prompting module, which leverages a vision-language model to generate learnable prompt tokens. These prompts explicitly preserve view-specific cues within the unified representation.
    
    \item We design a prompt-guided cross-view adapter, which injects the generated prompts into cross-view interaction to guide the learning of more discriminative view-specific feature representations.
    
    \item We propose a confidence-modulated mutual distillation loss to regularize the learning of the unified representation by dynamically suppressing noise propagation from less reliable views.
\end{itemize}
\section{Related Work}
\subsection{UAV-Ground Visual Tracking}

UGVT effectively alleviates the limitations of single-view tracking and enhances overall perception capability. Early studies mainly relied on task allocation or late-stage fusion across different views for collaborative surveillance. For example, Minaeian \textit{et al.}~\cite{UGVT1,UGVT2} used UAVs for coarse-grained detection and then employed unmanned ground vehicles for high-resolution localization. These approaches typically perform perception independently in each view and combine the results through geometric transformations or rule-based integration.

Recent UGVT studies have increasingly explored multi-view representation learning and achieved notable progress. For instance, Zhou \textit{et al.}~\cite{UGVT3} adopted graph neural networks to facilitate cross-view information fusion. Sun \textit{et al.}~\cite{MVCL} established a comprehensive UAV-ground tracking benchmark and proposed MVCL, which correlates and learns from feature cues extracted separately from each view. Similarly, Wang \textit{et al.}~\cite{uvcpnet} developed a collaborative perception framework in which features are extracted independently and then fused in a bird's-eye-view space using specialized collaboration modules. Hou \textit{et al.}~\cite{agc} proposed a multi-object tracking dataset for autonomous driving scenarios and provided solutions to occlusion challenges.

Although these methods have achieved significant progress, they still rely on separate feature extraction pipelines for different views and essentially perform tracking on each view individually. In contrast, our method adopts a unified framework for joint modeling and mutual supervision across both views. By leveraging unified token modeling together with visual-language-guided feature representation, the proposed method promotes more effective cross-view interaction while maintaining a compact overall framework.
\subsection{Multi-View Visual Tracking}

Current research on multi-view visual tracking mainly focuses on three categories: multi-UAV single-object tracking (SOT), multi-UAV multi-object tracking, and multi-camera object tracking.

In the area of multi-UAV SOT, Zhu \textit{et al.}~\cite{asnet} established the MDOT benchmark and proposed ASNet, which employs independent Siamese trackers with template sharing. However, this framework still adopts separate view branches and relies on relatively weak interaction between views. As a result, the target is essentially tracked independently in each view, and the complementary information across views is not fully exploited.

In the multi-UAV tracking setting, Liu \textit{et al.}~\cite{mianet} proposed MIA-Net, which uses homography-based geometric alignment to associate targets across different views. In addition, TransMDOT~\cite{transmdot} and CRM~\cite{CRM} leverage self-attention and view-invariant modules to enhance feature interaction. Nevertheless, these methods still follow a pipeline in which features are first extracted separately and then aligned or interacted afterward.

Similarly, multi-camera object tracking also suffers from isolated cross-view feature modeling. For example, Divotrack~\cite{divotrack} extends single-view trackers to the multi-view setting, but still processes each view independently. GMT~\cite{gmt} employs Transformers to model cross-view appearance and spatio-temporal features for similarity-based matching. More recently, MITracker~\cite{mitracker} performs multi-view bounding-box tracking through geometric projection and feature fusion, providing another approach to information integration across views. In general, these methods rely on independent view branches, where features are processed in parallel pipelines before later interaction. This design limits the full exploitation of complementary information across views.

In summary, most existing multi-view tracking frameworks rely on independent encoders and are characterized by weak interaction or late-stage fusion. As a result, they fail to achieve sufficiently deep cross-view interaction and lack explicit guidance for unified joint modeling.
\begin{figure*}[t]
  \centering
  \includegraphics[width=0.90\textwidth]{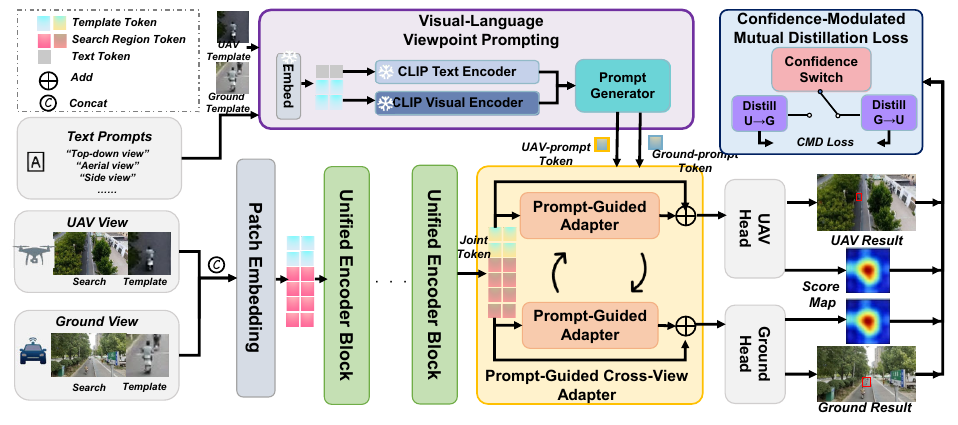}   
  \caption{The framework of the proposed VL-UniTrack method. Features of templates and search regions of UAV and ground view extracted by unified encoder are enhanced by the Prompt-Guided Cross-View Adapter. The PCVA is driven by viewpoint tokens generated via visual-language viewpoint prompting, which fuses text prompts with visual templates. The method is optimized using confidence-modulated mutual distillation loss, where a confidence switch determines the direction of distillation to mitigate error propagation from unreliable view.}   
  \label{fig:pipeline}  
\end{figure*}        
\begin{figure}[t]
\centering
  \includegraphics[width=\columnwidth]{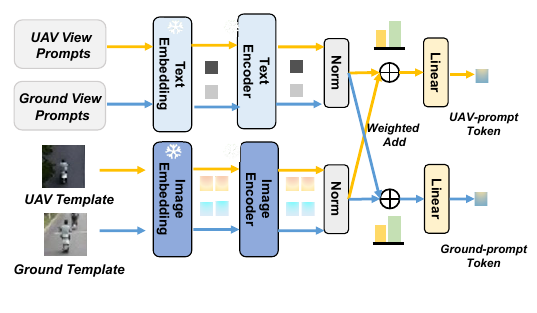}
  \caption{Detailed design of the visual-language viewpoint prompting module, which generates view-specific prompt tokens by fusing textual viewpoint priors with template visual features.} 
  \label{fig:VLGP}  
\end{figure}
\begin{figure}[t]
\centering
  \includegraphics[width=\columnwidth]{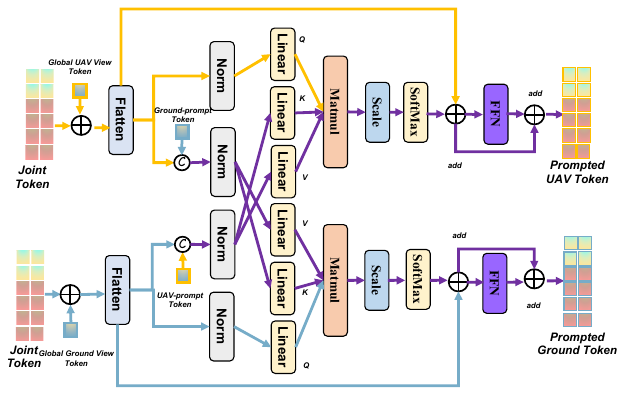}
  \caption{Detailed design of the prompt-guided cross-view adapter, which injects viewpoint prompts into cross-view interaction to learn more discriminative view-specific representations.} 
  \label{fig:PCVA}  
\end{figure}
\subsection{Single-View visual Tracking}

Single-view visual tracking can generally be categorized into classification-based, Siamese-based, and Transformer-based methods. Classification-based trackers, such as MDNet~\cite{mdnet} and DiMP~\cite{dimp}, formulate tracking as an online foreground-background discrimination problem. Although these methods maintain strong robustness through continuous model updates, they often suffer from high computational cost. To improve efficiency, Siamese-based trackers, including SiamFC~\cite{siamfc} and LightTrack~\cite{lighttrack}, adopt a dual-branch framework to match features between templates and search regions. While these methods achieve real-time performance, they rely heavily on static templates, which limits their adaptability to significant appearance changes and long-term tracking scenarios.

More recently, Transformer-based methods have been introduced to enable stronger feature modeling. Representative approaches, such as OSTrack~\cite{ostrack} and SwinTrack~\cite{swintrack}, replace conventional encoders with Transformer architectures, enabling more comprehensive feature extraction and relation modeling. Furthermore, sequential trackers such as SeqTrack~\cite{seqtrackv2} and ARTrack~\cite{artrackv2} incorporate temporal cues to improve localization accuracy. Nevertheless, extending such single-view methods to multi-view tracking remains challenging, as it requires handling the complex interaction and heterogeneity across different views.

\section{Method}
\subsection{Overview}
The overall framework of VL-UniTrack is illustrated in Fig.~\ref{fig:pipeline}. Given the UAV template-search pair $\mathcal{I}_u = \{Z_u, X_u\}$ and the ground-view pair $\mathcal{I}_g = \{Z_g, X_g\}$, we first form paired target-centered inputs by channel-wise concatenation. The unified visual encoder jointly projects the paired patches and produces a single joint feature representation $F_{\mathrm{J}}$. The VLVP module employs a frozen CLIP encoder to generate the view-specific prompt tokens $P_u$ and $P_g$ from textual viewpoint descriptions and visual templates. Conditioned on these prompts, PCVA transforms $F_{\mathrm{J}}$ into the enhanced UAV- and ground-specific representations $\hat{F}_u$ and $\hat{F}_g$. Parallel prediction heads then output the score maps and bounding boxes for the two views. During training, CMD applies confidence-gap-gated directional supervision between the two enhanced view-specific representations.
\subsection{Unified Visual Encoder}
Traditional UAV-ground visual tracking methods typically adopt a dual-stream architecture with separate encoders for UAV and ground views. Such independent feature extraction suffers from a semantic gap between the two views, which hinders effective cross-view interaction. To address this limitation, we employ a unified visual encoder~\cite{fastitpn} to project UAV and ground-view features into a shared semantic space. This design facilitates more thorough cross-view interaction and provides a solid foundation for subsequent learning of view-specific representations.

Given target-centered UAV and ground-view template crops $Z_u$ and $Z_g$, and their search-region crops $X_u$ and $X_g$, we resize the crops in each pair to the same spatial resolution and form $\bar{Z}=[Z_u;Z_g]_c$ and $\bar{X}=[X_u;X_g]_c$, where $[\cdot;\cdot]_c$ denotes channel-wise concatenation. Each spatial patch of $\bar{Z}$ or $\bar{X}$ therefore contains six channels formed from the two target-centered views at the same normalized index.

The six-channel joint projection first fuses local patches at the same spatial index in the two views to produce intermediate features:
\begin{equation}
    \begin{aligned}
    \mathbf{h}^{z}_{k} &= \mathbf{W}_{6}
    [\mathbf{z}_{u,k};\mathbf{z}_{g,k}]+\mathbf{b},\\
    \mathbf{h}^{x}_{k} &= \mathbf{W}_{6}
    [\mathbf{x}_{u,k};\mathbf{x}_{g,k}]+\mathbf{b},
    \end{aligned}
    \label{eq:joint_patch_projection}
\end{equation}
where $\mathbf{z}_{u,k},\mathbf{z}_{g,k}$ and $\mathbf{x}_{u,k},\mathbf{x}_{g,k}$ are flattened three-channel template and search patches at position $k$, respectively; $p$ is the initial patch size; $\mathbf{W}_{6}\in\mathbb{R}^{D_0\times 6p^2}$ and $\mathbf{b}\in\mathbb{R}^{D_0}$ are the joint projection weights and bias; and $\mathbf{h}^{z}_{k},\mathbf{h}^{x}_{k}\in\mathbb{R}^{D_0}$ are intermediate features, with $D_0$ denoting the initial projection's output channel dimension. The superscripts $z$ and $x$ distinguish the template and search regions, respectively.

Let $\mathcal{P}_{6}(\cdot)$ denote joint tokenization, comprising the six-channel projection, the existing hierarchical front-end processing of Fast-iTPN, and spatial unfolding. Its outputs satisfy $\mathcal{P}_{6}(\bar{Z})\in\mathbb{R}^{N_z\times D}$ and $\mathcal{P}_{6}(\bar{X})\in\mathbb{R}^{N_x\times D}$, where $N_z,N_x$ are the template and search token counts after front-end processing, and $D$ is the dimension of each token, with one token per row. We concatenate the search and template tokens along the sequence dimension and add positional embeddings:
\begin{equation}
    E_{\mathrm{J}} =
    \left[ \mathcal{P}_{6}(\bar{X}) \, ; \, \mathcal{P}_{6}(\bar{Z}) \right]
    + \Pi
    \in\mathbb{R}^{(N_x+N_z)\times D},
    \label{eq:input_embed}
\end{equation}
where $E_{\mathrm{J}}$ is the joint input sequence, the upright subscript $\mathrm{J}$ denotes a joint representation, $\Pi\in\mathbb{R}^{(N_x+N_z)\times D}$ is the positional embedding matrix, and $[\cdot;\cdot]$ denotes concatenation along the token dimension. Joint tokenization couples information from the two views within the tokens. The search and template tokens form a single joint sequence of length $N_x+N_z$, which is passed to the subsequent encoding stages:
\begin{equation}
    F_{\mathrm{J}} = \Phi(E_{\mathrm{J}}),
    \label{eq:backbone_out}
\end{equation}
where $\Phi(\cdot)$ denotes the subsequent Fast-iTPN encoding and search-feature readout. The readout extracts the $N_x$ search-region tokens from the encoded sequence and reshapes them into the joint feature map $F_{\mathrm{J}}\in\mathbb{R}^{D\times H\times W}$, with spatial dimensions $H,W$ satisfying $N_x=HW$. PCVA then adapts this joint feature map into view-specific representations.

\subsection{Visual-Language Viewpoint Prompting}
\label{sec:vlvp}

The unified encoder produces a compact joint representation in which information from the paired UAV and ground-view inputs has already been coupled. Although unified encoding reduces feature isolation, viewpoint-specific cues may still be insufficiently represented within the unified representation. We therefore introduce VLVP to provide explicit UAV and ground-view semantic guidance for deriving view-conditioned representations from $F_{\mathrm{J}}$. By leveraging the knowledge of CLIP~\cite{clip}, VLVP generates semantic tokens that serve as viewpoint prompts and preserve the viewpoint-specific characteristics required for robust tracking.

\label{par:language_role}
The visual branch in VLVP extracts target-specific appearance context from the initial target templates, while the language branch provides generic semantics for the UAV and ground viewpoints. Their fusion produces prompts that encode both target appearance and viewpoint information to guide cross-view interaction and view-specific representation learning. In our tracking setting, the target is specified by its initial bounding boxes and visual templates, and language descriptions provide viewpoint guidance.\label{par:language_role_end}

As illustrated in Fig.~\ref{fig:VLGP}, VLVP has parallel text and vision branches. To capture view-specific cues, the text branch averages CLIP embeddings of viewpoint descriptions:
\begin{equation}
    P_v^{\mathrm{L}} = \frac{1}{|\mathcal{T}_v|} \sum_{t \in \mathcal{T}_v} \Phi_{\mathrm{L}}(t), \quad v \in \{u, g\},
    \label{eq:text_prior}
\end{equation}
where $\Phi_{\mathrm{L}}(\cdot)$ denotes the language branch implemented using the frozen CLIP text encoder, $\mathcal{T}_v$ is the set of descriptions for view $v$, $t\in\mathcal{T}_v$ is an individual description, and $P_v^{\mathrm{L}}$ is the language prompt feature before fusion. The UAV descriptions are ``Top-down view'', ``Aerial view'' and ``Overhead shot'', while the ground-view descriptions are ``Side view'', ``Front view'' and ``Street level view''. The vision branch encodes the initial templates to extract target appearance information:
\begin{equation}
    P_v^{\mathrm{V}} = \Phi_{\mathrm{V}}(Z_v), \quad v \in \{u, g\},
    \label{eq:vis_ctx}
\end{equation}
where $\Phi_{\mathrm{V}}(\cdot)$ denotes the visual branch implemented using the frozen CLIP image encoder, $Z_v$ is the initial template image for view $v$, and $P_v^{\mathrm{V}}$ is the visual prompt feature before fusion. Weighted fusion and linear projection produce the final prompt tokens:
\begin{equation}
    P_v = \varphi \left( \mathrm{Norm}(P_v^{\mathrm{L}}) + \alpha \cdot \mathrm{Norm}(P_v^{\mathrm{V}}) \right), \quad v \in \{u, g\},
    \label{eq:prompt_gen}
\end{equation}
where $\mathrm{Norm}(\cdot)$ denotes L2 normalization, $\alpha$ is a learnable balancing parameter, and $\varphi(\cdot)$ projects features linearly into the tracker's embedding dimension $D$. The tokens $P_u$ and $P_g$ jointly encode viewpoint semantics and visual context to guide subsequent cross-view interaction. The features $P_v^{\mathrm{L}},P_v^{\mathrm{V}}$ are the pre-fusion prompt features within VLVP, while $P_v$ is the final fused prompt passed to PCVA.
Unlike a static view-identity embedding that only indicates whether a branch corresponds to the UAV or ground view, the proposed prompts jointly encode high-level viewpoint semantics and target-specific visual context extracted from the initial templates. They therefore provide instance-conditioned guidance for subsequent cross-view interaction.
\subsection{Prompt-Guided Cross-View Adapter}
To effectively exploit the complementary information between UAV and ground views, we propose a PCVA. This module is designed to establish viewpoint-aware interaction across the two views. Specifically, the viewpoint prompts generated by the VLVP module are incorporated into a cross-attention mechanism, so that cross-view interaction is guided by view-specific cues rather than relying solely on visual similarity. In PCVA, Cross-attention establishes the basic information exchange between the two views, the global view tokens provide branch identity cues, and the VLVP prompts introduce explicit viewpoint-aware visual-language guidance into the interaction.

Fig.~\ref{fig:PCVA} illustrates the internal prompt-guided interaction of PCVA. The module takes the joint feature tensor $F_{\mathrm{J}}$ as input and processes it through two symmetric branches. First, to adapt the unified features to each specific view, we inject learnable global view tokens through element-wise addition:
\begin{equation}
    F_u = \Psi(F_{\mathrm{J}} + \mathcal{G}_u),
\end{equation}
\begin{equation}
    F_g = \Psi(F_{\mathrm{J}} + \mathcal{G}_g),
\end{equation}
where $\mathcal{G}_u,\mathcal{G}_g$ are learnable global view tokens; $\Psi(\cdot)$ denotes spatial unfolding, converting a $D\times H\times W$ feature map into an $HW\times D$ token matrix in spatial row-major order, with each row representing the $D$-dimensional feature at one spatial location; and $F_u,F_g$ are the UAV and ground base features, respectively.

These base features are then refined through cross-attention-based interaction. Taking the UAV branch as an example, the goal is to enhance the UAV representation by querying complementary information from the ground view. To impose viewpoint guidance, we concatenate the VLVP-generated UAV prompt with the ground base features:
\begin{equation}
    K_u = V_u = [P_u; F_g],
    \label{eq:kv_construct}
\end{equation}
where $[\cdot;\cdot]$ denotes concatenation along the token dimension, $P_u$ is the fused UAV prompt, $F_g$ is the ground base feature, and $K_u,V_u$ are the key/value projection inputs. The concatenated sequence is layer-normalized after prompt insertion, while the query is normalized separately. We then apply the attention and feed-forward residual sublayers and integrate the adapter output through learnable residual scaling:
\begin{equation}
    \hat{F}_u = \gamma_u\,\mathrm{FFN}\!\left(\mathrm{MHCA}(F_u,K_u,V_u)\right) + F_u,
    \label{eq:final_out}
\end{equation}
where the three inputs of $\mathrm{MHCA}(\cdot,\cdot,\cdot)$ correspond to query, key, and value, respectively; $\mathrm{MHCA}(\cdot,\cdot,\cdot)$ and $\mathrm{FFN}(\cdot)$ are pre-normalized multi-head cross-attention and feed-forward residual sublayers with the internal skip connections in Fig.~\ref{fig:PCVA}; $\gamma_u,\gamma_g$ are branch-specific learnable parameters; and $\hat{F}_u$ is the enhanced UAV feature. Symmetrically, the ground branch queries with $F_g$ and constructs key/value inputs by concatenating $P_g$ with $F_u$ to obtain the enhanced ground feature $\hat{F}_g$.
\subsection{Confidence-Modulated Mutual Distillation Loss}
CMD lets the relatively more reliable view guide the other without forcing teacher assignment when their reliability is indistinguishable. It uses the maximum target-center response as an online relative reliability proxy:
\begin{equation}
    c_v=\max_{i,j}\mathbf{R}_v(i,j), \qquad v\in\{u,g\},
    \label{eq:conf_score}
\end{equation}
where $\mathbf{R}_v\in[0,1]^{H\times W}$ is the view-$v$ probability response map, $H,W$ are its spatial dimensions, $(i,j)$ indexes locations, and $c_v$ is the maximum response. This proxy represents the response strength at the most likely target location and requires no additional quality-estimation network. Importantly, CMD uses it only to compare the relative reliability of the two views rather than interpreting an individual response as an absolute probability of correct tracking.

During training, we perform target-region-weighted pooling on the view-specific PCVA features using the ground-truth Gaussian center heatmaps:
\begin{equation}
    \mathbf{f}_v=
    \frac{\sum_{i,j}\mathbf{M}_v(i,j)\hat{F}_v(i,j)}
    {\sum_{i,j}\mathbf{M}_v(i,j)+\epsilon},
    \qquad v\in\{u,g\},
    \label{eq:target_pooling}
\end{equation}
where $\mathbf{M}_v$ is the Gaussian center heatmap, $\hat{F}_v(i,j)$ the enhanced feature vector at the corresponding spatial location, $\mathbf{f}_v$ the pooled target feature, and $\epsilon$ a small positive constant for numerical stability. The pooled feature allows CMD to regularize target-relevant representations instead of directly matching the complete spatial feature maps.

We activate directional distillation only when the confidence gap exceeds a margin. The CMD objective is defined as
\begin{equation}
    \begin{aligned}
    \mathcal{L}_{\mathrm{cmd}}={}&
    \mathbb{I}(c_u>c_g+\delta)
    \mathcal{L}_{\cos}\!\left(\mathbf{f}_g,\mathrm{sg}[\mathbf{f}_u]\right)\\
    &+\mathbb{I}(c_g>c_u+\delta)
    \mathcal{L}_{\cos}\!\left(\mathbf{f}_u,\mathrm{sg}[\mathbf{f}_g]\right),
    \end{aligned}
    \label{eq:cmd_loss}
\end{equation}
where $\mathbb{I}(\cdot)$ is the indicator function, equal to 1 when the condition holds and 0 otherwise, $\mathrm{sg}[\cdot]$ denotes stop-gradient, and $\delta=0.05$ is the confidence-gap margin. The cosine loss is defined as
\begin{equation}
    \mathcal{L}_{\cos}(\mathbf{r}_1,\mathbf{r}_2)
    =1-\frac{\mathbf{r}_1^{\top}\mathbf{r}_2}
    {\|\mathbf{r}_1\|_2\|\mathbf{r}_2\|_2},
    \label{eq:cosine_loss}
\end{equation}
where $\mathbf{r}_1$ and $\mathbf{r}_2$ are the input feature vectors and $\|\cdot\|_2$ denotes the Euclidean norm. Thus, the higher-confidence view serves as a fixed teacher only when the relative difference is sufficiently clear. When $|c_u-c_g|\leq\delta$, neither term is activated and $\mathcal{L}_{\mathrm{cmd}}=0$. This confidence-gap gating filters samples with an ambiguous teacher direction, although it does not assume that every activated teacher is necessarily correct.

\begin{algorithm}[htbp]
\caption{Confidence-Modulated Mutual Distillation (CMD) loss}
\label{alg:cmd}
\begin{algorithmic}[1]
\Require Enhanced features $\hat{F}_u,\hat{F}_g$; probability maps $\mathbf{R}_u,\mathbf{R}_g$; Gaussian maps $\mathbf{M}_u,\mathbf{M}_g$; margin $\delta$
\Ensure Distillation loss $\mathcal{L}_{\mathrm{cmd}}$
\State $c_u\gets\max(\mathbf{R}_u)$; $c_g\gets\max(\mathbf{R}_g)$
\State $\mathbf{f}_u\gets\mathrm{WeightedPool}(\hat{F}_u,\mathbf{M}_u)$
\State $\mathbf{f}_g\gets\mathrm{WeightedPool}(\hat{F}_g,\mathbf{M}_g)$
\State $\mathcal{L}_{\mathrm{cmd}}\gets 0$
\If{$c_u>c_g+\delta$}
    \State $\mathcal{L}_{\mathrm{cmd}}\gets\mathcal{L}_{\cos}(\mathbf{f}_g,\mathrm{sg}[\mathbf{f}_u])$
\ElsIf{$c_g>c_u+\delta$}
    \State $\mathcal{L}_{\mathrm{cmd}}\gets\mathcal{L}_{\cos}(\mathbf{f}_u,\mathrm{sg}[\mathbf{f}_g])$
\EndIf
\State \Return $\mathcal{L}_{\mathrm{cmd}}$
\end{algorithmic}
\end{algorithm}

Following existing center-based trackers~\cite{ostrack}, each view-specific prediction head contains a target-center score-map branch and a bounding-box regression branch. The score-map branch is supervised by a focal loss computed against a Gaussian center heatmap, while the predicted bounding box is optimized using L1 and generalized IoU losses. The overall objective further incorporates the proposed CMD loss:
\begin{equation}
    \begin{aligned}
    \mathcal{L}
    ={}&\sum_{v\in\{u,g\}}
    \left(
    \lambda_{\mathrm{focal}}\mathcal{L}_{\mathrm{focal}}^v
    +\lambda_1\mathcal{L}_1^v
    +\lambda_{\mathrm{giou}}\mathcal{L}_{\mathrm{giou}}^v
    \right)\\
    &+\lambda_{\mathrm{cmd}}\mathcal{L}_{\mathrm{cmd}},
    \end{aligned}
    \label{eq:total_loss}
\end{equation}
where $\mathcal{L}$ is the total loss, $\mathcal{L}_{\mathrm{focal}}^v$, $\mathcal{L}_1^v$ and $\mathcal{L}_{\mathrm{giou}}^v$ are the focal, L1, and GIoU losses for view $v$, respectively, and $\lambda_{\mathrm{focal}}$, $\lambda_1$, $\lambda_{\mathrm{giou}}$ and $\lambda_{\mathrm{cmd}}$ are loss weights set to $1$, $5$, $2$ and $1$, respectively, in all experiments.
\begin{figure}[t]
\centering
  \includegraphics[width=\columnwidth]{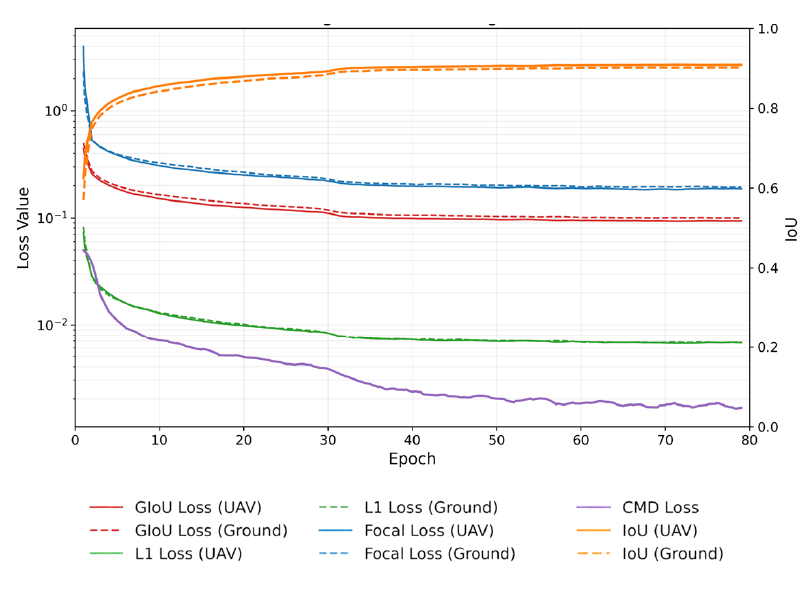}
  \caption{Visualization results of Optimization Trajectory.} 
  \label{fig:visualloss}  
\end{figure}
\begin{figure*}[t]
\centering
  \includegraphics[width=0.70\textwidth]{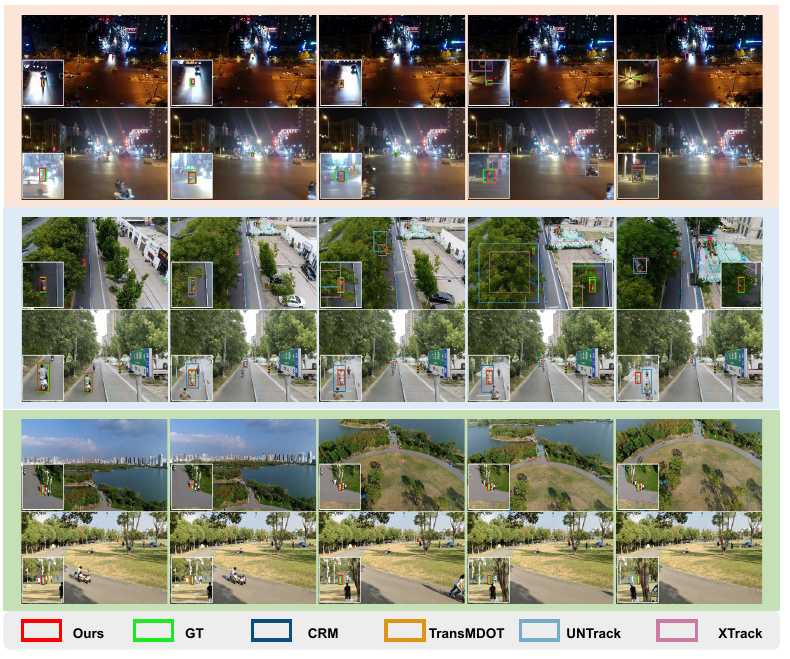}
  \caption{Qualitative comparison between the proposed method and state-of-the-art methods on the UGVT test set. The ground truth is marked in green.} 
  \label{fig:visual}  
\end{figure*}
\section{Experiments}
\subsection{Dataset and Evaluation}
We evaluate the proposed method on the UGVT dataset~\cite{MVCL}, which is currently the largest benchmark for UAV-ground visual tracking. It contains over 200 pairs of synchronized UAV and ground-view tracking sequences, covering diverse target categories such as pedestrians and vehicles. Moreover, the dataset includes a wide range of challenging scenarios, including severe occlusion, similar objects, camera motion, and large scale variation. The main challenge of UGVT lies in the drastic viewpoint discrepancy between UAV and ground views, which leads to significant appearance divergence for the same target and imposes stringent requirements on the tracker’s ability to maintain robust cross-view consistency.

Following prior work, we adopt two evaluation metrics: Precision Rate (PR) and Success Rate (SR).

PR measures center-location accuracy at the 20-pixel threshold and is averaged over all test sequences:
\begin{equation}
  \mathrm{PR} =
  \frac{1}{|\mathcal{Q}|} \sum_{q\in\mathcal{Q}}
  \frac{1}{N_q} \sum_{i=1}^{N_q}
  \mathbb{I}(\mathrm{CLE}_{q,i}\leq20),
  \label{eq:precision_rate}
\end{equation}
where $\mathcal{Q}$ is the test-sequence set, $N_q$ is the frame count of sequence $q$, and $\mathrm{CLE}_{q,i}$ is the Euclidean distance between predicted and ground-truth centers in its $i$-th frame, in pixels; $\mathbb{I}(\cdot)$ is the indicator function defined above.

For the $q$-th sequence, the success rate at IoU threshold $\tau$ is $S_q(\tau)=\frac{1}{N_q}\sum_{i=1}^{N_q}\mathbb{I}(\operatorname{IoU}_{q,i}>\tau)$. SR is reported as the area under the success curve, implemented as the mean success rate over 21 IoU thresholds:
\begin{equation}
  \mathrm{SR}
  =\frac{1}{|\mathcal{Q}|}\sum_{q\in\mathcal{Q}}
  \frac{1}{|\Theta|}\sum_{\tau\in\Theta} S_q(\tau),
  \label{eq:success_rate}
\end{equation}
where $\Theta=\{0,0.05,\ldots,1\}$ denotes the set of IoU thresholds, $\operatorname{IoU}_{q,i}$ is the intersection over union between the predicted and ground-truth boxes in frame $i$ of sequence $q$, and $S_q(\tau)$ is the success rate of that sequence at threshold $\tau$; $\mathcal{Q}$ and $N_q$ have the same meanings as in Eq.~\eqref{eq:precision_rate}.
\subsection{Implementation Details}
The proposed method is trained on a server equipped with a CPU clocked at 5.2~GHz and a GPU with 32~GB of memory. The model is trained for 80 epochs with a batch size of 8, and each epoch contains 10,000 samples. We use AdamW with a base learning rate of $1\times10^{-4}$ and an encoder learning rate of $1\times10^{-5}$, corresponding to an encoder multiplier of 0.1. The learning rate is decayed by a factor of 0.1 every 30 epochs, and the weight decay is set to $1\times10^{-4}$. All other training parameters are consistent with the baseline~\cite{sutrack}. The model has 98.67M trainable parameters and achieves real-time performance at approximately 30 FPS.

\begin{table}[t]
    \centering
    \scriptsize
    \setlength{\tabcolsep}{3.2pt}
    \renewcommand{\arraystretch}{1.12} %
    \caption{Comparison between the proposed method and state-of-the-art trackers on UGVT test set. The best results are highlighted in \textbf{bold}. The performance is evaluated in terms of PR and SR. $^{*}$ denotes multi-view tracking methods.}
    \label{tab:benchmark}
    \resizebox{\columnwidth}{!}{%
    \begin{tabular}{c|c|cc|cc|cc|c}
        \toprule
        \multirow{2}{*}{Method} & \multirow{2}{*}{Year} &
        \multicolumn{2}{c|}{Ground View} &
        \multicolumn{2}{c|}{UAV View} &
        \multicolumn{2}{c|}{Average} &
        \multirow{2}{*}{FPS$\uparrow$} \\
        & & SR$\uparrow$ & PR$\uparrow$ & SR$\uparrow$ & PR$\uparrow$ & SR$\uparrow$ & PR$\uparrow$ & \\
        \midrule
        DiMP50     & ICCV19  & 0.567 & 0.815 & 0.606 & 0.793 & 0.587 & 0.804 & 35 \\
        SiamFC++   & AAAI20  & 0.466 & 0.675 & 0.534 & 0.703 & 0.500 & 0.689 & 57 \\
        TransT     & CVPR21  & 0.578 & 0.773 & 0.572 & 0.728 & 0.575 & 0.751 & 43 \\
        KeepTrack  & ICCV21  & 0.599 & 0.818 & 0.636 & 0.821 & 0.618 & 0.820 & 16 \\
        SparseTT   & IJCAI22 & 0.622 & 0.813 & 0.585 & 0.716 & 0.604 & 0.765 & 36 \\
        MixFormer  & CVPR22  & 0.536 & 0.689 & 0.592 & 0.772 & 0.564 & 0.731 & 28 \\
        ToMP50     & CVPR22  & 0.621 & 0.837 & 0.645 & 0.835 & 0.633 & 0.836 & 24 \\
        OSTrack    & ECCV22  & 0.580 & 0.758 & 0.572 & 0.740 & 0.576 & 0.749 & \textbf{58} \\
        AiATrack   & ECCV22  & 0.590 & 0.781 & 0.629 & 0.822 & 0.610 & 0.802 & 40 \\
        CTTrack    & AAAI23  & 0.601 & 0.812 & 0.612 & 0.819 & 0.607 & 0.816 & 19 \\
        MVCL$^{*}$ & TCSVT23 & 0.637 & 0.844 & 0.623 & 0.786 & 0.630 & 0.815 & 21 \\
        TransMDOT$^{*}$ & TCSVT23 & 0.587 & 0.733 & 0.687 & 0.853 & 0.637 & 0.793 & 30 \\
        UNTrack    & CVPR24  & 0.654 & 0.797 & 0.709 & 0.858 & 0.682 & 0.828 & 15 \\
        CRM$^{*}$ & TCSVT24 & 0.408 & 0.586 & 0.623 & 0.840 & 0.515 & 0.713 & 38 \\
        XTrack     & ICCV25  & 0.666 & 0.851 & 0.700 & 0.868 & 0.683 & 0.859 & 17 \\
        SSTrack    & AAAI25  & 0.612 & 0.766 & 0.702 & 0.874 & 0.657 & 0.820 & 33 \\
        SuTrack    & AAAI25  & 0.587 & 0.745 & 0.649 & 0.818 & 0.618 & 0.781 & 40 \\
        MITracker$^{*}$ & CVPR25 & 0.549 & 0.694 & 0.641 & 0.798 & 0.595 & 0.746 & 16 \\
        \midrule
        Ours       &         & \textbf{0.704} & \textbf{0.853} & \textbf{0.717} & \textbf{0.886} &\textbf{ 0.711} & \textbf{0.869} & 30 \\
        \bottomrule
    \end{tabular}%
    }
\end{table}
\begin{table}[t]
    \centering
    \scriptsize
    \setlength{\tabcolsep}{2.5pt}
    \renewcommand{\arraystretch}{1.15}
    \caption{Component analysis on UGVT test set. PCVA represents prompt-guided cross-view adapter, CMD represents confidence-modulated mutual distillation loss, and VLVP represents visual-language viewpoint prompting. $\Delta\,(\%)$ denotes the improvement over the baseline in the first row. The best results are highlighted in \textbf{bold}.}
    \label{tab:ablation}
    \begin{tabular}{ccc|cccc|cccc}
        \toprule
            \multirow{2}{*}{PCVA} &\multirow{2}{*}{CMD}&\multirow{2}{*}{VLVP} & \multicolumn{4}{c|}{UAV View} & \multicolumn{4}{c}{Ground View} \\
        &&&PR$\uparrow$ & $\Delta(\%)$ & SR$\uparrow$ & $\Delta(\%)$ &
        PR$\uparrow$ & $\Delta(\%)$ & SR$\uparrow$ & $\Delta(\%)$ \\
        \midrule
             &      &      & 0.818 & --    & 0.649 & --    & 0.745 & -- & 0.587 & -- \\
        \ding{51} &      &      & 0.842 & +2.4  & 0.681 & +3.2  & 0.829 & +8.4  & 0.684 & +9.7  \\
             & \ding{51} &      & 0.841 & +2.3   & 0.683 & +3.4   & 0.820 & +7.5  & 0.679 & +9.2  \\
        \ding{51} & \ding{51} &      & 0.859 & +4.1   & 0.700 & +5.1   & 0.833 & +8.8  & 0.690 & +10.3  \\
        \ding{51} & \ding{51} & \ding{51} & \textbf{0.886} & +6.8 & \textbf{0.717} & +6.8 &\textbf{0.853} & +10.8  & \textbf{0.704} & +11.7  \\
        \bottomrule
    \end{tabular}
\end{table}

\begin{table}[htbp]
    \centering
    \scriptsize
    \setlength{\tabcolsep}{2.5pt}
    \renewcommand{\arraystretch}{1.15}
    \caption{Ablation study of VLVP module on UGVT test set. The best results are highlighted in \textbf{bold}. $\Delta\,(\%)$ denotes the improvement over w/o VLVP.}
    \label{tab:ablationvlgp}
    \begin{tabular}{c|cccc|cccc}
        \toprule
            \multirow{2}{*}{Setting} & \multicolumn{4}{c|}{UAV View} & \multicolumn{4}{c}{Ground View} \\
        &PR$\uparrow$ & $\Delta(\%)$ & SR$\uparrow$ & $\Delta(\%)$ &
        PR$\uparrow$ & $\Delta(\%)$ & SR$\uparrow$ & $\Delta(\%)$ \\
        \midrule
            w/o VLVP       & 0.859 & --    & 0.700 & --    & 0.833 & -- & 0.690 & -- \\
            Learnable Embedding & 0.869 & +1.0 & 0.706 & +0.6 & 0.846 & +1.3 & 0.699 & +0.9 \\
            Only Visual Prompt  & 0.878 & +1.9 & 0.712 & +1.2 & 0.850 & +1.7 & 0.699 & +0.9 \\
             Only Text Prompt   & 0.869 & +1.0 & 0.705 & +0.5 & 0.841 & +0.8 & 0.695 & +0.5 \\
             Ours & \textbf{0.886} & +2.7 & \textbf{0.717} & +1.7 & \textbf{0.853} & +2.0 & \textbf{0.704} & +1.4 \\
 
        \bottomrule
    \end{tabular}
\end{table}
\begin{figure}[t]
\centering
  \includegraphics[width=\columnwidth]{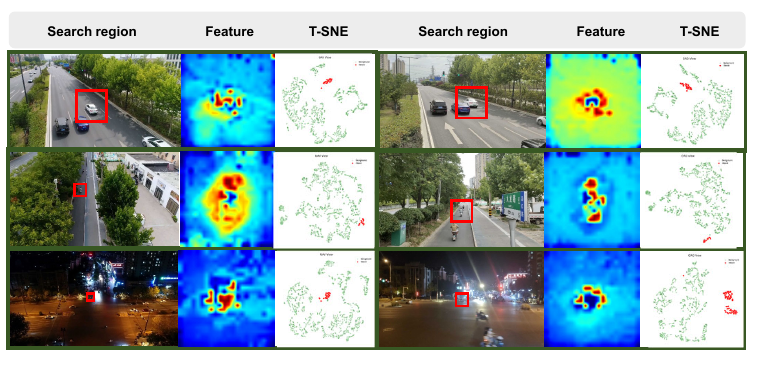}
  \caption{Feature visualization and t-SNE results of the proposed method.} 
  \label{fig:visuafeat}  
\end{figure}
\subsection{Comparison with State-of-the-Arts Method}
To comprehensively evaluate the effectiveness of the proposed method, we compare VL-UniTrack with a broad range of state-of-the-art trackers on the UGVT test set, including DiMP50~\cite{dimp}, SiamFC++~\cite{siamfc++}, TransT~\cite{transt}, KeepTrack~\cite{keeptrack}, SparseTT~\cite{sparsett}, MixFormer~\cite{mixformerv2}, ToMP50~\cite{tomp}, OSTrack~\cite{ostrack}, AiATrack~\cite{aiatrack}, CTTrack~\cite{CTTtrack}, MVCL~\cite{MVCL}, TransMDOT~\cite{transmdot}, CRM~\cite{CRM}, MITracker~\cite{mitracker}, UNTrack~\cite{untrack}, XTrack~\cite{xtrack}, SSTrack~\cite{SSTrack}, and SuTrack~\cite{sutrack}. The quantitative results are summarized in Table~\ref{tab:benchmark}. As shown in the table, VL-UniTrack achieves the best performance on both the ground and UAV views, as well as on the overall average metrics, while still maintaining real-time inference at 30 FPS. These results demonstrate that the proposed unified framework can effectively exploit complementary information across the two views without sacrificing practical efficiency.

More specifically, on the ground view, VL-UniTrack outperforms the strongest competing method, XTrack, improving PR from 0.851 to 0.853 and SR from 0.666 to 0.704, corresponding to gains of 0.2\% and 3.8\%, respectively. On the UAV view, our method improves the best competing SR result from 0.709 to 0.717, and the best competing PR result from 0.874 to 0.886, yielding gains of 0.8\% and 1.2\%, respectively. In terms of the overall average performance, VL-UniTrack achieves gains over XTrack of 1.0\% in PR and 2.8\% in SR. These consistent improvements across both views indicate that our method does not favor one branch at the expense of the other; instead, it enables UAV and ground-view features to reinforce each other within a unified representation space, leading to more stable and reliable tracking under challenging cross-view conditions.

To further strengthen the experimental comparison directly related to UAV-ground tracking, we supplement the existing MVCL baseline with three representative multi-view trackers: TransMDOT, CRM, and MITracker. As shown in Table~\ref{tab:benchmark}, VL-UniTrack achieves the highest PR and SR on both views among these multi-view baselines. Compared with TransMDOT, which performs best on average among the three added methods, VL-UniTrack improves the average PR from 0.793 to 0.869 and the average SR from 0.637 to 0.711, corresponding to gains of 7.6\% and 7.4\%, respectively. These results further demonstrate the effectiveness of our method for collaborative UAV-ground tracking.

\subsection{Ablation Study}
To thoroughly validate the contribution of each component in the proposed framework, we conduct a series of component-wise ablation experiments, and the results are summarized in Table~\ref{tab:ablation}.

The first row reports the baseline performance: 0.818 and 0.649 in UAV PR and SR, respectively, and 0.745 and 0.587 in ground-view PR and SR, respectively. After introducing the PCVA module, the model obtains gains of 2.4\% and 3.2\% on the UAV view, together with larger improvements of 8.4\% and 9.7\% on the ground view. This indicates that prompt-guided cross-view interaction is particularly beneficial when the target is more difficult to localize. Adding CMD alone on top of the baseline also consistently improves performance, yielding gains of 2.3\% and 3.4\% on the UAV view, as well as 7.5\% and 9.2\% on the ground view. These results verify that confidence-aware mutual distillation can effectively enhance cross-view collaboration during training.

When PCVA and CMD are used together, the performance is further improved, with gains of 4.1\% and 5.1\% on the UAV view and 8.8\% and 10.3\% on the ground view, indicating that the two modules are complementary. Finally, after incorporating VLVP, the full model achieves the best performance, with overall improvements of 6.8\% and 6.8\% on the UAV view and 10.8\% and 11.7\% on the ground view over the baseline. These results clearly demonstrate that each module contributes positively to the final performance, and that their combination yields the best performance among the evaluated configurations.

\subsection{Targeted Analysis of Prompt Guidance in PCVA}
To further isolate the roles of different components in PCVA, we progressively introduce Cross-attention, the global view tokens, and the VLVP prompts. As shown in Table~\ref{tab:pcva_prompt_guidance}, introducing Cross-attention without prompt information improves PR from 0.841 to 0.853 and SR from 0.683 to 0.698 on the UAV view, and PR from 0.820 to 0.832 and SR from 0.679 to 0.692 on the ground view. The resulting average PR and SR are 0.843 and 0.695, respectively. This result shows that basic cross-view attention can effectively exploit complementary information between the two views. Adding the global view tokens yields 0.846 and 0.695 in average PR and SR, respectively, corresponding to comparatively small further changes. After incorporating the VLVP prompts, the model achieves the best average PR and SR of 0.869 and 0.711, respectively. Compared with Cross-attention without prompt information, the complete prompt-guided design achieves improvements in average PR and SR of 2.6\% and 1.6\%, respectively. These results show that Cross-attention establishes basic cross-view information exchange, the global view tokens provide branch identity cues, and the VLVP prompts further provide effective viewpoint-aware visual-language guidance for the interaction.

\begin{table}[t]
\centering
\scriptsize
\setlength{\tabcolsep}{3pt}
\renewcommand{\arraystretch}{1.15}
\caption{Targeted ablation of prompt guidance in PCVA on the UGVT test set. The best results are in \textbf{bold}.}
\label{tab:pcva_prompt_guidance}
\begin{tabular}{l|cc|cc|cc}
\toprule
\multirow{2}{*}{Setting} & \multicolumn{2}{c|}{UAV View} & \multicolumn{2}{c|}{Ground View} & \multicolumn{2}{c}{Average} \\
& PR$\uparrow$ & SR$\uparrow$ & PR$\uparrow$ & SR$\uparrow$ & PR$\uparrow$ & SR$\uparrow$ \\
\midrule
CMD only / w/o PCVA & 0.841 & 0.683 & 0.820 & 0.679 & 0.830 & 0.681 \\
$+$ Cross-attention & 0.853 & 0.698 & 0.832 & 0.692 & 0.843 & 0.695 \\
$+$ Global view token & 0.859 & 0.700 & 0.833 & 0.690 & 0.846 & 0.695 \\
$+$ VLVP prompt & \textbf{0.886} & \textbf{0.717} & \textbf{0.853} & \textbf{0.704} & \textbf{0.869} & \textbf{0.711} \\
\bottomrule
\end{tabular}
\end{table}

\subsection{Encoder Architecture Analysis}
To distinguish the effect of parameter sharing from unified representation learning, we compare three encoder architectures. Independent Two-stream employs two encoders with independent parameters and processes the UAV and ground views separately. Shared-weight Two-stream uses the same encoder weights for both views but still performs two separate forward passes, and therefore does not construct joint tokens or joint features inside the encoder. Our Unified Joint Encoder concatenates the paired target-centered inputs channel-wise, projects every paired patch into one view-coupled token through a six-channel joint patch embedding, and encodes the resulting compact joint sequence in a single Fast-iTPN forward pass. All three architectures retain the same VLVP, PCVA, CMD, and prediction heads, so the comparison focuses on the encoder design.

\begin{table}[t]
\centering
\scriptsize
\setlength{\tabcolsep}{2.5pt}
\renewcommand{\arraystretch}{1.15}
\caption{Comparison of encoder architectures on UGVT. Online FLOPs and latency are model-forward costs; the best results are in \textbf{bold}.}
\label{tab:encoder_architecture}
\begin{tabular}{l|cc|cc}
\toprule
\multirow{2}{*}{Architecture} & \multicolumn{2}{c|}{Average} & \multicolumn{2}{c}{Efficiency} \\
& PR$\uparrow$ & SR$\uparrow$ & FLOPs$\downarrow$ & Latency$\downarrow$ \\
\midrule
Independent Two-stream & 0.825 & 0.684 & 410.6G & 14.48 ms \\
Shared-weight Two-stream & 0.838 & 0.691 & 410.6G & 14.46 ms \\
Unified Joint Encoder (Ours) & \textbf{0.869} & \textbf{0.711} & \textbf{213.8}G & \textbf{7.98} ms \\
\bottomrule
\end{tabular}
\end{table}

As shown in Table~\ref{tab:encoder_architecture}, Independent Two-stream obtains 0.825 and 0.684 in average PR and SR, respectively. Sharing the encoder weights improves the average PR and SR to 0.838 and 0.691, respectively, showing that parameter sharing itself provides a certain benefit. However, because the two views are still encoded in separate forward passes, the online computation and latency remain almost unchanged. More importantly, our Unified Joint Encoder achieves 0.869 and 0.711 in average PR and SR, respectively, corresponding to gains of 3.1\% and 2.0\% over Shared-weight Two-stream. It also reduces Online FLOPs from 410.6G to 213.8G and Online Latency from 14.46 ms to 7.98 ms. These results show that the benefit is not attributable to parameter sharing alone, but also arises from the compact joint representation formed by paired-view joint patch projection. Compared with Independent Two-stream, our method achieves improvements in average PR and SR of 4.4\% and 2.7\%, respectively, while requiring substantially lower online computation, thereby achieving a better accuracy-efficiency balance.

\subsection{Confidence Reliability Analysis}
To examine whether the maximum score-map response used by CMD provides a reliable cue for teacher selection, we compare it with normalized entropy and average peak-to-correlation energy (APCE). Each measure is evaluated by whether it selects the view with higher IoU, and Meaningful-gap Accuracy is additionally reported for pairs satisfying $|\Delta\mathrm{IoU}|>0.05$. Since the three measures have different numerical scales, they are compared at matched activation coverage.

\begin{table}[t]
\centering
\scriptsize
\setlength{\tabcolsep}{7pt}
\renewcommand{\arraystretch}{1.15}
\caption{Comparison of score-map confidence measures for CMD teacher selection at matched activation coverage.}
\label{tab:confidence_reliability}
\begin{tabular}{lcc}
\toprule
Confidence measure & Switch Acc. & Meaningful-gap Acc. \\
\midrule
Normalized entropy & 49.26\% & 52.30\% \\
APCE & 64.31\% & 72.70\% \\
Maximum response (Ours) & \textbf{65.98}\% & \textbf{74.47}\% \\
\bottomrule
\end{tabular}
\end{table}

As shown in Table~\ref{tab:confidence_reliability}, maximum response achieves the highest teacher-selection accuracy of 65.98\%, compared with 49.26\% for normalized entropy and 64.31\% for APCE. Its Meaningful-gap Accuracy reaches 74.47\%, also exceeding normalized entropy (52.30\%) and APCE (72.70\%). These results verify that maximum response provides an effective and lightweight relative reliability cue for CMD. Moreover, confidence-gap gating with $\delta=0.05$ improves teacher-selection accuracy from 62.86\% to 65.98\% and Meaningful-gap Accuracy from 69.97\% to 74.47\%, demonstrating the benefit of filtering samples with an ambiguous teacher direction.

\subsection{Ablation Study of VLVP}
\label{sec:vlvp_ablation}
\subsubsection{Overall Ablation Analysis}
\label{sec:vlvp_overall}
To investigate the respective contributions of visual and textual prompts within the VLVP module, we conduct a detailed ablation study, and the results are reported in Table~\ref{tab:ablationvlgp}. First, removing the VLVP module entirely leads to a clear performance drop, which confirms the necessity of guiding the learning of view-specific feature representations. Second, using only visual prompts yields considerable improvements, suggesting that visual context plays an important role in fine-grained appearance alignment. Similarly, using only textual prompts also brings consistent gains, indicating that semantic viewpoint cues can effectively guide feature representation learning. Most importantly, the full model, which combines both visual and textual prompts, achieves the best performance across all metrics. This result demonstrates that detailed appearance information from visual features and high-level semantic context from textual prompts are complementary, thereby enabling more robust tracking under drastic view changes.

\subsubsection{Comparison with Learnable Embedding}
\label{sec:learnable_embedding}
To further distinguish the complete VLVP from a simple view-identity encoding, we add a Learnable Embedding baseline that injects static UAV/ground view embeddings into PCVA through the same prompt interface, without CLIP semantics or target-specific visual context. As shown in Table~\ref{tab:ablationvlgp}, Learnable Embedding obtains 0.869 and 0.706 in UAV PR and SR, respectively, compared with 0.886 and 0.717 for the complete VLVP. On the ground view, it obtains 0.846 and 0.699 in PR and SR, respectively, compared with 0.853 and 0.704 for the complete VLVP. Compared with Learnable Embedding, the complete VLVP achieves improvements of 1.7\% and 1.1\% in UAV PR and SR, respectively, and 0.7\% and 0.5\% in ground-view PR and SR, respectively. Together with the Only Visual Prompt and Only Text Prompt results, this shows that viewpoint-semantic priors and target-specific visual context provide complementary guidance beyond static view identity.\label{par:learnable_embedding_end}

\subsubsection{Sensitivity to Fusion Weight}
In VLVP, the fusion weight $\alpha$ is a learnable scalar that is optimized jointly with the network, rather than a manually tuned hyperparameter. To explicitly examine its sensitivity, we additionally fix $\alpha$ to three representative values and evaluate the resulting models on the UGVT test set. As shown in Table~\ref{tab:alpha_sensitivity}, the learned setting achieves 0.869 and 0.711 in average PR and SR, respectively, outperforming all fixed alternatives. These results indicate that learning $\alpha$ allows VLVP to adaptively balance viewpoint semantics and visual context, yielding robust performance without manual tuning.

\begin{table}[t]
\centering
\scriptsize
\setlength{\tabcolsep}{6pt}
\renewcommand{\arraystretch}{1.15}
\caption{Sensitivity analysis of the VLVP fusion weight $\alpha$ on the UGVT test set. ``Learned'' denotes the default setting, in which $\alpha$ is optimized jointly with the model.}
\label{tab:alpha_sensitivity}
\begin{tabular}{c|cc|cc|cc}
\toprule
\multirow{2}{*}{$\alpha$} & \multicolumn{2}{c|}{UAV View} & \multicolumn{2}{c|}{Ground View} & \multicolumn{2}{c}{Average} \\
& PR$\uparrow$ & SR$\uparrow$ & PR$\uparrow$ & SR$\uparrow$ & PR$\uparrow$ & SR$\uparrow$ \\
\midrule
0.25 & 0.849 & 0.694 & 0.847 & 0.703 & 0.848 & 0.699 \\
Learned & \textbf{0.886} & \textbf{0.717} & \textbf{0.853} & \textbf{0.704} & \textbf{0.869} & \textbf{0.711} \\
0.75 & 0.852 & 0.697 & 0.849 & 0.697 & 0.850 & 0.697 \\
1.00 & 0.859 & 0.702 & 0.834 & 0.690 & 0.847 & 0.696 \\
\bottomrule
\end{tabular}
\end{table}
\subsubsection{Effect of Viewpoint-Specific Language Guidance}
\label{sec:prompt_semantics}
To further verify the role of viewpoint-aware language prompts, we compare View-specific, Generic, and Swapped settings. In the Generic setting, both UAV and ground views use the prompt ``an image of an object''. In the Swapped setting, the original UAV and ground-view prompts are directly exchanged.
\begin{table}[H]
\centering
\scriptsize
\setlength{\tabcolsep}{5pt}
\renewcommand{\arraystretch}{1.15}
\caption{Ablation results of different language prompt settings on UGVT.}
\label{tab:prompt_semantics}
\begin{tabular}{l|cc|cc|cc}
\toprule
\multirow{2}{*}{Prompt setting} & \multicolumn{2}{c|}{UAV View} & \multicolumn{2}{c|}{Ground View} & \multicolumn{2}{c}{Average} \\
& PR$\uparrow$ & SR$\uparrow$ & PR$\uparrow$ & SR$\uparrow$ & PR$\uparrow$ & SR$\uparrow$ \\
\midrule
View-specific & \textbf{0.886} & \textbf{0.717} & \textbf{0.853} & 0.704 & \textbf{0.869} & \textbf{0.711} \\
Generic & 0.863 & 0.700 & 0.814 & 0.677 & 0.839 & 0.688 \\
Swapped & 0.852 & 0.698 & 0.850 & \textbf{0.706} & 0.851 & 0.702 \\
\bottomrule
\end{tabular}
\end{table}
\label{par:prompt_semantics_analysis}
As shown in Table~\ref{tab:prompt_semantics}, View-specific achieves the best average PR and SR. Compared with Generic, it achieves improvements of 3.0\% and 2.3\% in average PR and SR, respectively; compared with Swapped, the corresponding improvements are 1.8\% and 0.9\%. Generic removes the explicit viewpoint distinction from the language prompts, whereas Swapped disrupts the correspondence between the viewpoint semantics and their respective branches. These results show that viewpoint-specific language semantics and their correct branch correspondence contribute to tracking performance.\label{par:prompt_semantics_end}
\begin{figure}[!t]
\centering
  \includegraphics[width=\columnwidth]{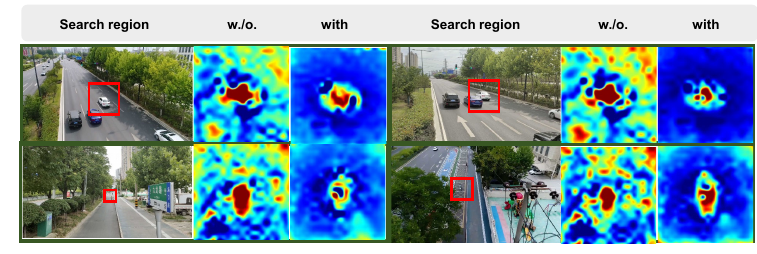}
  \caption{Feature maps inside PCVA with and without prompts.} 
  \label{fig:feataba}  
\end{figure}

\subsection{Efficiency Analysis}
\begin{table}[H]
\centering
\scriptsize
\setlength{\tabcolsep}{2pt}
\renewcommand{\arraystretch}{1.15}
\caption{Efficiency comparison under the same inference setting. Parameters are reported as trainable/total, while computation and latency are reported as initialization/online costs.}
\label{tab:flops}
\begin{tabular}{c|cccc}
\toprule
\multirow{2}{*}{Method} & Params & FLOPs & Latency & \multirow{2}{*}{Peak Mem.} \\
& Train./Total & Init./Online & Init./Online & \\
\midrule
Baseline & 89.49M/89.49M & --/208.1G & --/7.05 ms & 433.2 MB \\
Ours & 98.67M/249.95M & 17.5G/213.8G & 4.10/7.98 ms & 790.4 MB \\
\bottomrule
\end{tabular}
\end{table}
To comprehensively evaluate the computational efficiency of VL-UniTrack, we further report the parameter count, FLOPs, model forward latency, and peak GPU memory in Table~\ref{tab:flops}. The increase in peak memory from 433.2 MB to 790.4 MB mainly arises from the frozen CLIP ViT-B/32 model used to extract visual prompts. Although CLIP introduces 151.28M frozen parameters, the number of trainable parameters increases only from 89.49M to 98.67M, corresponding to an increase of 10.26\%. Meanwhile, the online computation increases only from 208.1 GFLOPs to 213.8 GFLOPs, and the model forward latency increases from 7.05 ms to 7.98 ms. These results show that the additional memory is primarily associated with the frozen CLIP weights, while the increases in trainable model size, online computation, and inference latency remain limited. Online latency is measured using CUDA Events for model forward only after caching the initial CLIP prompts; it excludes image loading, crop generation, CPU preprocessing, host-to-device transfer, post-processing, and result writing, whereas the FPS reported in Table~\ref{tab:benchmark} includes the complete tracking pipeline.
\subsection{Attribute-Level Evaluation}
To further evaluate the robustness of VL-UniTrack under different challenging conditions, we conduct an attribute-level evaluation on the two challenge attributes provided with the released UGVT data, namely Full Occlusion and Out-of-View. All trackers are evaluated under the same one-pass evaluation protocol using PR and SR.
\begin{table}[H]
\centering
\scriptsize
\setlength{\tabcolsep}{6pt}
\renewcommand{\arraystretch}{1.15}
\caption{Performance comparison under different challenge attributes on UGVT. Results are reported as PR/SR; the best results are in \textbf{bold}.}
\label{tab:attribute_evaluation}
\begin{tabular}{l|cc}
\toprule
Tracker & Full Occlusion & Out-of-View \\
\midrule
UNTrack & 0.663 / 0.500 & 0.883 / 0.665 \\
XTrack & 0.671 / 0.464 & 0.888 / 0.643 \\
SSTrack & 0.655 / 0.487 & 0.872 / 0.696 \\
CRM$^{*}$ & 0.583 / 0.395 & 0.597 / 0.451 \\
TransMDOT$^{*}$ & 0.635 / 0.472 & 0.881 / 0.714 \\
MITracker$^{*}$ & 0.713 / 0.510 & 0.825 / 0.636 \\
VL-UniTrack & \textbf{0.768} / \textbf{0.552} & \textbf{0.957} / \textbf{0.798} \\
\bottomrule
\end{tabular}
\end{table}
As shown in Table~\ref{tab:attribute_evaluation}, VL-UniTrack achieves the best PR and SR under both attributes. It obtains 0.768 and 0.552 in PR and SR, respectively, under Full Occlusion, achieving improvements over the strongest competing PR and SR of 5.5\% and 4.2\%, respectively. Under Out-of-View, it achieves 0.957 and 0.798 in PR and SR, respectively. When one view lacks reliable target observations, VL-UniTrack exploits complementary information from the other view to maintain stable tracking, demonstrating robustness to single-view failure.
\subsection{Visualization Results}

\textbf{Tracking Result Visualization.}
To provide a more intuitive demonstration of the effectiveness of the proposed method, we visualize qualitative comparisons with the representative multi-view trackers CRM and TransMDOT, together with UNTrack and XTrack, as shown in Fig.~\ref{fig:visual}. Across challenging scenarios involving occlusion, small targets, and background interference, our method maintains accurate target localization in both views and produces more stable tracking results than the compared methods. These qualitative comparisons further demonstrate the effectiveness of VL-UniTrack in exploiting cross-view complementarity for robust UAV-ground visual tracking.

\textbf{Feature Visualization.}
To further demonstrate the effectiveness of our approach, we visualize the intermediate features during tracking and perform a t-SNE analysis, as shown in Fig.~\ref{fig:visuafeat}. It can be observed that, throughout the tracking process, even when the target appears extremely small, the responses of our method from both views remain consistently focused on the target. Moreover, the t-SNE results show that our method can clearly separate foreground targets from background regions. These analyses provide additional evidence that our method learns discriminative and view-consistent feature representations, thereby contributing to robust and stable tracking performance.

\textbf{Visualization of the Role of Prompts in PCVA.}
To provide more direct evidence of the role of prompts in PCVA, we visualize the intermediate feature maps inside PCVA with and without prompts, as shown in Fig.~\ref{fig:feataba}. Without prompt guidance, the feature responses are more diffuse and are easily distracted by surrounding background structures. After introducing the prompts, the responses become more compact and concentrate more accurately on the target regions, with substantially reduced activation on irrelevant areas. This phenomenon is consistently observed across different scenes and in both UAV and ground views. These visualizations support our claim that the proposed prompts explicitly guide feature interaction in PCVA and help the network learn more discriminative view-aware representations.
\subsection{Optimization Trajectory}

To validate the stability of the proposed optimization strategy, we visualize the evolution of individual loss components together with the tracking performance (IoU) during training. As illustrated in Fig.~\ref{fig:visualloss}, all loss terms, including the focal score-map loss, GIoU loss, L1 loss, and the proposed CMD loss, exhibit a consistent and steady downward trend. Notably, the convergence of the CMD loss indicates that mutual distillation effectively regularizes the unified feature space without disrupting the learning of the primary tracking tasks. Correspondingly, as the losses decrease, the IoU scores for both UAV and ground views consistently improve and eventually stabilize at a high level. These results confirm that the proposed multi-objective optimization formulation leads to robust tracking performance and ensures a stable training process.
\subsection{Tracking Stability and Robustness Analysis}
\begin{figure}[t]
\centering
  \includegraphics[width=0.47\textwidth]{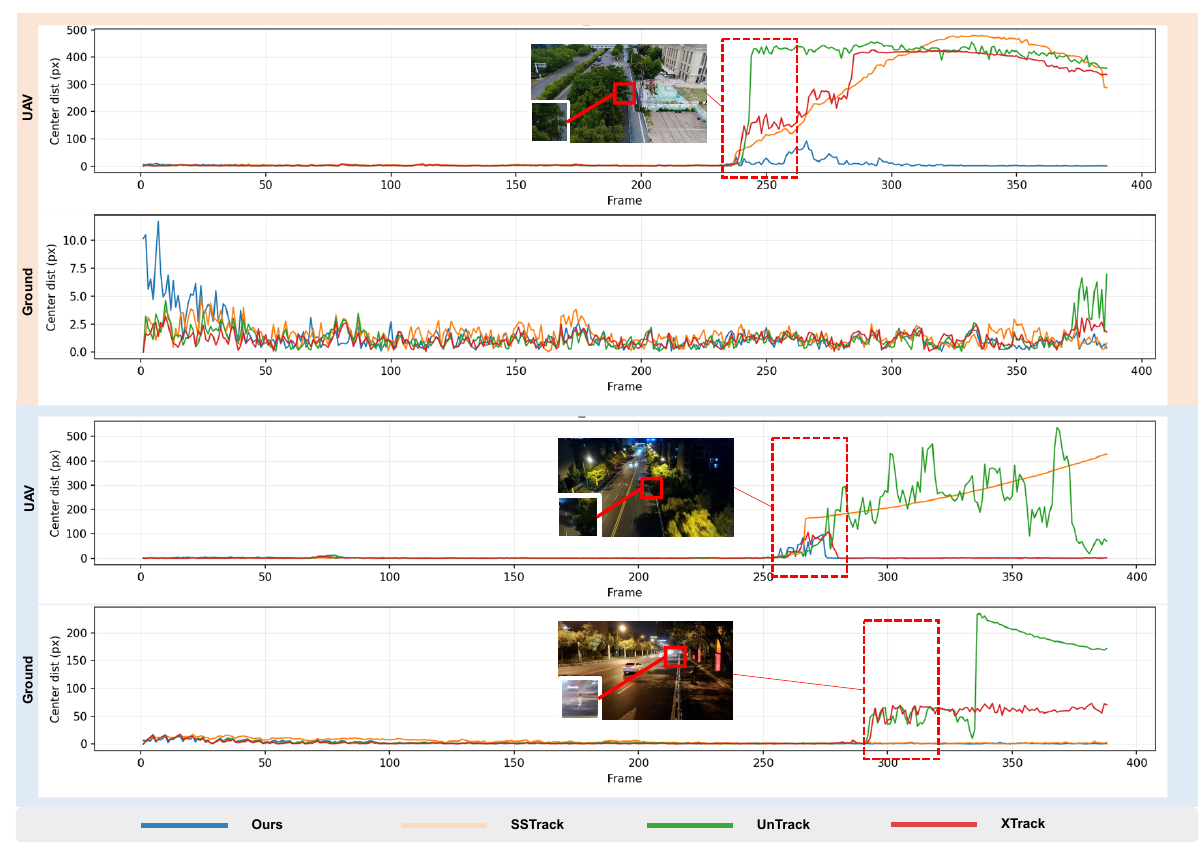}   
  \caption{Visualization of Center Location Error (CLE) on challenging sequences, which displays the frame-by-frame error evolution of different trackers. The curves represent the Euclidean distance (in pixels) between the center of the predicted bounding box and the ground truth. A lower value indicates smaller location deviation and higher tracking accuracy.} 
  \label{fig:visualstable}  
\end{figure}
\begin{figure}[!b]
\centering
  \includegraphics[width=0.48\textwidth,pagebox=cropbox]{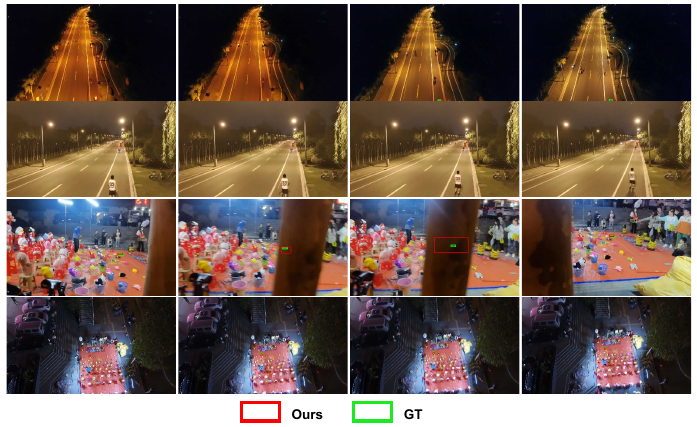}
  \caption{Representative failure cases and recovery processes of VL-UniTrack under simultaneous cross-view degradation. Each pair of rows shows two synchronized views of the same sequence, and columns correspond to successive frames. Red and green boxes denote our predictions and the ground truth, respectively.}
  \label{fig:failurecase}
\end{figure}
To further validate the robustness of VL-UniTrack, we select two challenging sequences and visualize the frame-level tracking deviation using the Center Location Error (CLE) metric. CLE is defined as the Euclidean distance between the predicted center and the ground-truth center, where a lower value indicates higher tracking accuracy.

As illustrated in Fig.~\ref{fig:visualstable}, when the target undergoes severe occlusion or is distracted by similar objects in one view, as reflected by the spikes in the error curves, competing methods often exhibit large deviations or even tracking failure. In contrast, VL-UniTrack consistently maintains a low and stable error curve throughout the sequences. This stability can be attributed to the cross-view complementarity enabled by PCVA: when one view is degraded, the network can effectively exploit reliable visual cues from the unobstructed counterpart to correct the prediction. These results demonstrate the effectiveness of cross-view collaboration in handling asymmetric degradation across views.
\subsection{Failure Case Analysis}

To provide a more comprehensive view of both the strengths and limitations of VL-UniTrack, we further analyze representative failure cases in Fig.~\ref{fig:failurecase}. Under single-view degradation, such as an extremely small target, severe occlusion, or the target moving outside the field of view, the other view can retain a reliable observation, maintain stable localization, and assist the degraded view in recovering once its visual evidence returns. Temporary drift may occur in the more extreme case where both views simultaneously provide insufficient evidence, for example, when the target is extremely small in one view while being severely occluded or outside the field of view in the other. Once either view regains reliable target evidence, cross-view collaboration can correct the prediction and resume stable dual-view tracking. These cases demonstrate the stability and recovery capability of VL-UniTrack under single-view degradation, while the simultaneous loss of reliable observations in both views remains a challenging condition.
\subsection{Discussion}
\label{sec:language_discussion}
The unified framework enables visual features from the UAV and ground views to interact sufficiently within a shared representation space, while viewpoint-specific cues also need to be fully represented. VLVP combines target appearance information from the initial templates with language-based viewpoint semantics to guide view-specific representations. The complete VLVP achieves higher average PR and SR than the Only Visual Prompt, Only Text Prompt and Learnable Embedding settings in Table~\ref{tab:ablationvlgp}, as well as the Generic setting in Table~\ref{tab:prompt_semantics}. These comparisons show that our combination of viewpoint semantics and target appearance information provides the most effective guidance among these alternatives, further improving tracking performance within the unified framework. In our tracking setting, the target is specified by its initial bounding boxes and visual templates, while language provides viewpoint-level semantic guidance.

Specifying targets through language descriptions is another valuable capability. Representative studies, including Tracking by Natural Language Specification by Li \textit{et al.}~\cite{li2017language} and Siamese Natural Language Tracker~\cite{snlt}, demonstrate the potential of language for expressing target semantics, supporting natural interaction, and complementing visual cues. Meanwhile, incorporating this capability into our framework requires consideration of potential inconsistencies between fixed object descriptions and dynamic visual observations: as the target's appearance or relative position changes, the original description may become inaccurate and interfere with localization. In future work, we will explore combining language-driven target specification with temporal visual evidence to improve the flexibility of target selection while maintaining stable UAV-ground collaborative tracking.\label{par:language_discussion_end}

\section{Conclusion}
In this paper, we proposed VL-UniTrack, a unified framework for UAV-ground visual tracking. Specifically, we constructed a unified joint encoder to process dual-view inputs jointly, thereby reducing feature isolation between the two views and enabling sufficient cross-view interaction. To address the representation ambiguity caused by implicit appearance matching, we designed a visual-language viewpoint prompting module that fuses language descriptions with visual features to generate viewpoint-aware prompts. These prompts are further utilized by the prompt-guided cross-view adapter to explicitly guide cross-view feature interaction and the learning of view-specific representations. In addition, we introduced a confidence-modulated mutual distillation loss to stabilize training. Extensive experiments on the benchmark demonstrate that VL-UniTrack achieves state-of-the-art performance.

\bibliographystyle{IEEEbib}
\bibliography{mybib}

@String(AAAI = {AAAI})

@inproceedings{untrack,
  title={Single-Model and Any-Modality for Video Object Tracking},
  author={Wu, Zongwei and Zheng, Jilai and Ren, Xiangxuan and Vasluianu, Florin-Alexandru and Ma, Chao and Paudel, Danda Pani and Van Gool, Luc and Timofte, Radu},
  booktitle={Proceedings of the IEEE/CVF Conference on Computer Vision and Pattern Recognition},
  year={2024},
  pages={19156--19166}
}

@inproceedings{transt,
  title={Transformer Tracking},
  author={Chen, Xin and Yan, Bin and Zhu, Jiawen and Wang, Dong and Yang, Xiaoyun and Lu, Huchuan},
  booktitle={Proceedings of the IEEE/CVF Conference on Computer Vision and Pattern Recognition},
  pages={8126--8135},
  year={2021}
}

@inproceedings{ostrack,
  title={Joint Feature Learning and Relation Modeling for Tracking: A One-Stream Framework},
  author={Ye, Botao and Chang, Hong and Ma, Bingpeng and Shan, Shiguang and Chen, Xilin},
  booktitle={Proceedings of the European Conference on Computer Vision},
  year={2022},
  pages={341--357},
  doi={10.1007/978-3-031-20047-2_20}
}

@inproceedings{dimp,
  title={Learning Discriminative Model Prediction for Tracking},
  author={Bhat, Goutam and Danelljan, Martin and Van Gool, Luc and Timofte, Radu},
  booktitle={Proceedings of the IEEE/CVF International Conference on Computer Vision},
  year={2019},
  pages={6182--6191}
}

@misc{seqtrackv2,
  title={Unified Sequence-to-Sequence Learning for Single- and Multi-Modal Visual Object Tracking},
  author={Chen, Xin and Kang, Ben and Zhu, Jiawen and Wang, Dong and Peng, Houwen and Lu, Huchuan},
  year={2024},
  note={arXiv:2304.14394v3}
}

@inproceedings{mixformerv2,
  title={{MixFormerV2}: Efficient Fully Transformer Tracking},
  author={Cui, Yutao and Song, Tianhui and Wu, Gangshan and Wang, Limin},
  booktitle={Advances in Neural Information Processing Systems},
  year={2023},
  volume={36},
  pages={58736--58751}
}

@inproceedings{aiatrack,
  title={{AiATrack}: Attention in Attention for Transformer Visual Tracking},
  author={Gao, Shenyuan and Zhou, Chunluan and Ma, Chao and Wang, Xinggang and Yuan, Junsong},
  booktitle={Proceedings of the European Conference on Computer Vision},
  year={2022},
  pages={146--164}
}

@inproceedings{artrackv2,
  author={Bai, Yifan and Zhao, Zeyang and Gong, Yihong and Wei, Xing},
  title={{ARTrackV2}: Prompting Autoregressive Tracker Where to Look and How to Describe},
  booktitle={Proceedings of the IEEE/CVF Conference on Computer Vision and Pattern Recognition},
  year={2024},
  pages={19048--19057}
}

@inproceedings{mdnet,
  title={Learning Multi-Domain Convolutional Neural Networks for Visual Tracking},
  author={Nam, Hyeonseob and Han, Bohyung},
  booktitle={Proceedings of the IEEE Conference on Computer Vision and Pattern Recognition},
  year={2016},
  pages={4293--4302}
}

@inproceedings{siamfc,
  title={Fully-Convolutional Siamese Networks for Object Tracking},
  author={Bertinetto, Luca and Valmadre, Jack and Henriques, Jo{\~a}o F. and Vedaldi, Andrea and Torr, Philip H. S.},
  booktitle={Proceedings of the European Conference on Computer Vision Workshops},
  year={2016},
  pages={850--865},
  doi={10.1007/978-3-319-48881-3_56}
}

@inproceedings{siamfc++,
  title={{SiamFC++}: Towards Robust and Accurate Visual Tracking with Target Estimation Guidelines},
  author={Xu, Yinda and Wang, Zeyu and Li, Zuoxin and Yuan, Ye and Yu, Gang},
  booktitle={Proceedings of the AAAI Conference on Artificial Intelligence},
  year={2020},
  volume={34},
  number={7},
  pages={12549--12556},
  doi={10.1609/aaai.v34i07.6944}
}

@inproceedings{swintrack,
  title={{SwinTrack}: A Simple and Strong Baseline for Transformer Tracking},
  author={Lin, Liting and Fan, Heng and Zhang, Zhipeng and Xu, Yong and Ling, Haibin},
  booktitle={Advances in Neural Information Processing Systems},
  year={2022},
  volume={35},
  pages={16743--16754}
}

@inproceedings{sutrack,
  title={{SuTrack}: Towards Simple and Unified Single Object Tracking},
  author={Chen, Xin and Kang, Ben and Geng, Wanting and Zhu, Jiawen and Liu, Yi and Wang, Dong and Lu, Huchuan},
  booktitle={Proceedings of the AAAI Conference on Artificial Intelligence},
  volume={39},
  number={2},
  pages={2239--2247},
  year={2025}
}

@inproceedings{SSTrack,
  title={Decoupled Spatio-Temporal Consistency Learning for Self-Supervised Tracking},
  author={Zheng, Yaozong and Zhong, Bineng and Liang, Qihua and Li, Ning and Song, Shuxiang},
  booktitle={Proceedings of the AAAI Conference on Artificial Intelligence},
  volume={39},
  number={10},
  pages={10635--10643},
  year={2025}
}

@inproceedings{xtrack,
  title={{XTrack}: Multimodal Training Boosts {RGB-X} Video Object Trackers},
  author={Tan, Yuedong and Wu, Zongwei and Fu, Yuqian and Zhou, Zhuyun and Sun, Guolei and Zamfir, Eduard and Ma, Chao and Paudel, Danda and Van Gool, Luc and Timofte, Radu},
  booktitle={Proceedings of the IEEE/CVF International Conference on Computer Vision},
  year={2025},
  pages={5734--5744}
}

@inproceedings{lighttrack,
  author={Yan, Bin and Peng, Houwen and Wu, Kan and Wang, Dong and Fu, Jianlong and Lu, Huchuan},
  title={{LightTrack}: Finding Lightweight Neural Networks for Object Tracking via One-Shot Architecture Search},
  booktitle={Proceedings of the IEEE/CVF Conference on Computer Vision and Pattern Recognition},
  year={2021},
  pages={15180--15189}
}

@inproceedings{UGVT1,
  title={Crowd Detection and Localization Using a Team of Cooperative {UAV}/{UGV}s},
  author={Minaeian, Sara and Liu, Jian and Son, Young-Jun},
  booktitle={Proceedings of the IIE Annual Conference and Expo},
  pages={595--604},
  year={2015}
}

@article{UGVT2,
  title={Vision-Based Target Detection and Localization via a Team of Cooperative {UAV} and {UGV}s},
  author={Minaeian, Sara and Liu, Jian and Son, Young-Jun},
  journal={IEEE Transactions on Systems, Man, and Cybernetics: Systems},
  volume={46},
  number={7},
  pages={1005--1016},
  year={2016},
  doi={10.1109/TSMC.2015.2491878}
}

@article{UGVT3,
  title={Multi-Robot Collaborative Perception with Graph Neural Networks},
  author={Zhou, Yang and Xiao, Jiuhong and Zhou, Yue and Loianno, Giuseppe},
  journal={IEEE Robotics and Automation Letters},
  volume={7},
  number={2},
  pages={2289--2296},
  year={2022}
}

@article{MVCL,
  title={{UAV}-Ground Visual Tracking: A Unified Dataset and Collaborative Learning Approach},
  author={Sun, Dengdi and Cheng, Leilei and Chen, Song and Li, Chenglong and Xiao, Yun and Luo, Bin},
  journal={IEEE Transactions on Circuits and Systems for Video Technology},
  volume={34},
  number={5},
  pages={3619--3632},
  year={2024},
  publisher={IEEE},
  doi={10.1109/TCSVT.2023.3316990}
}

@article{uvcpnet,
  title={{AVCPNet}: An {AAV}-Vehicle Collaborative Perception Network for {3-D} Object Detection},
  author={Wang, Yuchao and Wang, Zhirui and Cheng, Peirui and Tian, Pengju and Yuan, Ziyang and Tian, Jing and Wang, Wensheng and Zhao, Liangjin},
  journal={IEEE Transactions on Geoscience and Remote Sensing},
  year={2025},
  publisher={IEEE},
  volume={63},
  pages={1--16},
  doi={10.1109/TGRS.2025.3546669}
}

@misc{agc,
  title={{AGC-Drive}: A Large-Scale Dataset for Real-World Aerial-Ground Collaboration in Driving Scenarios},
  author={Hou, Yunhao and Zou, Bochao and Zhang, Min and Chen, Ran and Yang, Shangdong and Zhang, Yanmei and Zhuo, Junbao and Chen, Siheng and Chen, Jiansheng and Ma, Huimin},
  year={2025},
  note={arXiv:2506.16371}
}

@article{divotrack,
  title={{DIVOTrack}: A Novel Dataset and Baseline Method for Cross-View Multi-Object Tracking in Diverse Open Scenes},
  author={Hao, Shengyu and Liu, Peiyuan and Zhan, Yibing and Jin, Kaixun and Liu, Zuozhu and Song, Mingli and Hwang, Jenq-Neng and Wang, Gaoang},
  journal={International Journal of Computer Vision},
  volume={132},
  number={4},
  pages={1075--1090},
  year={2024}
}

@article{asnet,
  title={Multi-Drone-Based Single Object Tracking with Agent Sharing Network},
  author={Zhu, Pengfei and Zheng, Jiayu and Du, Dawei and Wen, Longyin and Sun, Yiming and Hu, Qinghua},
  journal={IEEE Transactions on Circuits and Systems for Video Technology},
  volume={31},
  number={10},
  pages={4058--4070},
  year={2021},
  publisher={IEEE},
  doi={10.1109/TCSVT.2020.3045747}
}

@article{mianet,
  title={Robust Multi-Drone Multi-Target Tracking to Resolve Target Occlusion: A Benchmark},
  author={Liu, Zhihao and Shang, Yuanyuan and Li, Timing and Chen, Guanlin and Wang, Yu and Hu, Qinghua and Zhu, Pengfei},
  journal={IEEE Transactions on Multimedia},
  volume={25},
  pages={1462--1476},
  year={2023},
  publisher={IEEE}
}

@article{transmdot,
  title={Cross-Drone Transformer Network for Robust Single Object Tracking},
  author={Chen, Guanlin and Zhu, Pengfei and Cao, Bing and Wang, Xing and Hu, Qinghua},
  journal={IEEE Transactions on Circuits and Systems for Video Technology},
  volume={33},
  number={9},
  pages={4552--4563},
  year={2023},
  publisher={IEEE}
}

@article{CRM,
  title={Consistent Representation Mining for Multi-Drone Single Object Tracking},
  author={Xue, Yuanliang and Jin, Guodong and Shen, Tao and Tan, Lining and Wang, Nian and Gao, Jing and Wang, Lianfeng},
  journal={IEEE Transactions on Circuits and Systems for Video Technology},
  volume={34},
  number={11},
  pages={10845--10859},
  year={2024},
  publisher={IEEE}
}

@inproceedings{mitracker,
  author={Xu, Mengjie and Zhu, Yitao and Jiang, Haotian and Li, Jiaming and Shen, Zhenrong and Wang, Sheng and Huang, Haolin and Wang, Xinyu and Zhang, Han and Yang, Qing and Wang, Qian},
  title={{MITracker}: Multi-View Integration for Visual Object Tracking},
  booktitle={Proceedings of the IEEE/CVF Conference on Computer Vision and Pattern Recognition},
  year={2025},
  pages={27176--27185}
}

@misc{gmt,
  title={{GMT}: A Robust Global Association Model for Multi-Target Multi-Camera Tracking},
  author={Fan, Huijie and Zhao, Tinghui and Wang, Qiang and Fan, Baojie and Tang, Yandong and Liu, LianQing},
  year={2024},
  note={arXiv:2407.01007v1}
}

@inproceedings{sparsett,
  title={{SparseTT}: Visual Tracking with Sparse Transformers},
  author={Fu, Zhihong and Fu, Zehua and Liu, Qingjie and Cai, Wenrui and Wang, Yunhong},
  booktitle={Proceedings of the International Joint Conference on Artificial Intelligence},
  pages={905--912},
  year={2022},
  doi={10.24963/ijcai.2022/127}
}

@inproceedings{tomp,
  title={Transforming Model Prediction for Tracking},
  author={Mayer, Christoph and Danelljan, Martin and Bhat, Goutam and Paul, Matthieu and Paudel, Danda Pani and Yu, Fisher and Van Gool, Luc},
  booktitle={Proceedings of the IEEE/CVF Conference on Computer Vision and Pattern Recognition},
  pages={8731--8740},
  year={2022}
}

@inproceedings{keeptrack,
  title={Learning Target Candidate Association to Keep Track of What Not to Track},
  author={Mayer, Christoph and Danelljan, Martin and Paudel, Danda Pani and Van Gool, Luc},
  booktitle={Proceedings of the IEEE/CVF International Conference on Computer Vision},
  pages={13444--13454},
  year={2021}
}

@inproceedings{CTTtrack,
  title={Compact Transformer Tracker with Correlative Masked Modeling},
  author={Song, Zikai and Luo, Run and Yu, Junqing and Chen, Yi-Ping Phoebe and Yang, Wei},
  booktitle={Proceedings of the AAAI Conference on Artificial Intelligence},
  volume={37},
  number={2},
  pages={2321--2329},
  year={2023}
}

@article{fastitpn,
  title={{Fast-iTPN}: Integrally Pre-Trained Transformer Pyramid Network with Token Migration},
  author={Tian, Yunjie and Xie, Lingxi and Qiu, Jihao and Jiao, Jianbin and Wang, Yaowei and Tian, Qi and Ye, Qixiang},
  journal={IEEE Transactions on Pattern Analysis and Machine Intelligence},
  year={2024},
  publisher={IEEE},
  volume={46},
  number={12},
  pages={9766--9779},
  doi={10.1109/TPAMI.2024.3429508}
}

@inproceedings{clip,
  title={Learning Transferable Visual Models from Natural Language Supervision},
  author={Radford, Alec and Kim, Jong Wook and Hallacy, Chris and Ramesh, Aditya and Goh, Gabriel and Agarwal, Sandhini and Sastry, Girish and Askell, Amanda and Mishkin, Pamela and Clark, Jack and Krueger, Gretchen and Sutskever, Ilya},
  booktitle={Proceedings of the International Conference on Machine Learning},
  pages={8748--8763},
  year={2021}
}

@inproceedings{li2017language,
  title={Tracking by Natural Language Specification},
  author={Li, Zhenyang and Tao, Ran and Gavves, Efstratios and Snoek, Cees G. M. and Smeulders, Arnold W. M.},
  booktitle={Proceedings of the IEEE Conference on Computer Vision and Pattern Recognition},
  pages={6495--6503},
  year={2017}
}

@inproceedings{snlt,
  title={Siamese Natural Language Tracker: Tracking by Natural Language Descriptions with Siamese Trackers},
  author={Feng, Qi and Ablavsky, Vitaly and Bai, Qinxun and Sclaroff, Stan},
  booktitle={Proceedings of the IEEE/CVF Conference on Computer Vision and Pattern Recognition},
  pages={5851--5860},
  year={2021}
}

@article{UAVUGV1,
  title={A Comprehensive Review of {UAV}-{UGV} Collaboration: Advancements and Challenges},
  author={Munasinghe, Isuru and Perera, Asanka and Deo, Ravinesh C.},
  journal={Journal of Sensor and Actuator Networks},
  volume={13},
  number={6},
  year={2024},
  publisher={MDPI},
  eid={81},
  doi={10.3390/jsan13060081}
}

@article{UAVUGV2,
  title={A Hybrid Path Planning Method in Unmanned Air/Ground Vehicle ({UAV}/{UGV}) Cooperative Systems},
  author={Li, Jianqiang and Deng, Genqiang and Luo, Chengwen and Lin, Qiuzhen and Yan, Qiao and Ming, Zhong},
  journal={IEEE Transactions on Vehicular Technology},
  volume={65},
  number={12},
  pages={9585--9596},
  year={2016},
  publisher={IEEE}
}

\clearpage
\raggedend
\makeatletter
\def\@IEEEBIOskipN{1.5\baselineskip}
\makeatother

\begin{IEEEbiography}
[{\includegraphics[trim=40bp 0bp 40bp 0bp,clip,width=1in,height=1.25in,keepaspectratio]{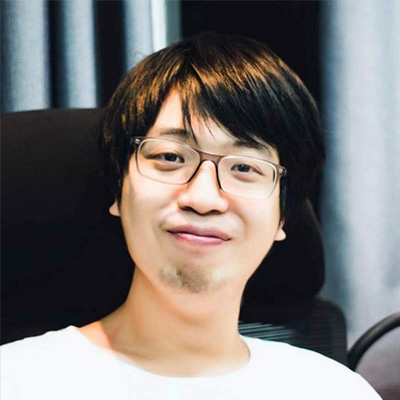}}]{Boyue Xu} (Student Member, IEEE) received the M.S. degree from the School of Software Engineering, Nanjing University, Nanjing, China. He is currently pursuing the Ph.D. degree with the Department of Computer Science and Technology, Nanjing University. His research interests include multi-modal representation learning, visual object tracking, and video object segmentation. He has published papers in journals and conferences, including Pattern Recognition, Image and Vision Computing, The Visual Computer, the IEEE International Conference on Multimedia and Expo, and the ACM International Conference on Multimedia Retrieval.
\end{IEEEbiography}

\begin{IEEEbiography}[{\includegraphics[trim=0bp 195bp 0bp 0bp,clip,width=1in,height=1.25in,keepaspectratio]{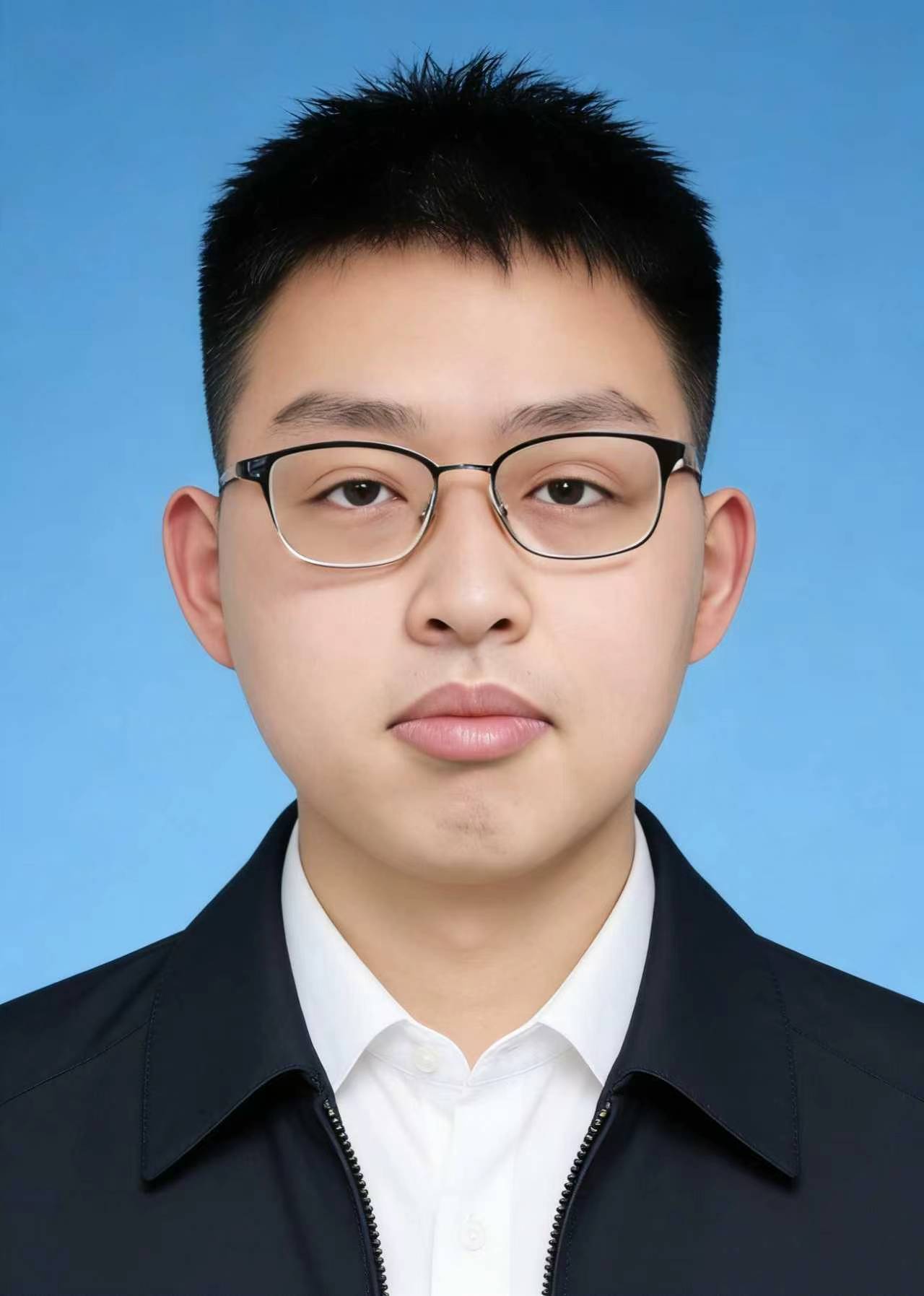}}]{Ruichao Hou} (Member, IEEE) received his Ph.D. degree from the Department of Computer Science and Technology, Nanjing University, in 2023. He is currently an Associate Professor with the School of Artificial Intelligence, China Pharmaceutical University. His research mainly focuses on multi-modal representation learning. He has published more than 40 papers in top-tier journals and conferences. He also serves as a reviewer for several prestigious journals, such as IEEE TPAMI, IEEE TNNLS, IEEE TMM, and IEEE TCSVT.\end{IEEEbiography}

\vfill
\newpage
 \begin{IEEEbiography}[{\includegraphics[trim=0bp 154bp 0bp 0bp,clip,width=1in,height=1.25in,keepaspectratio]{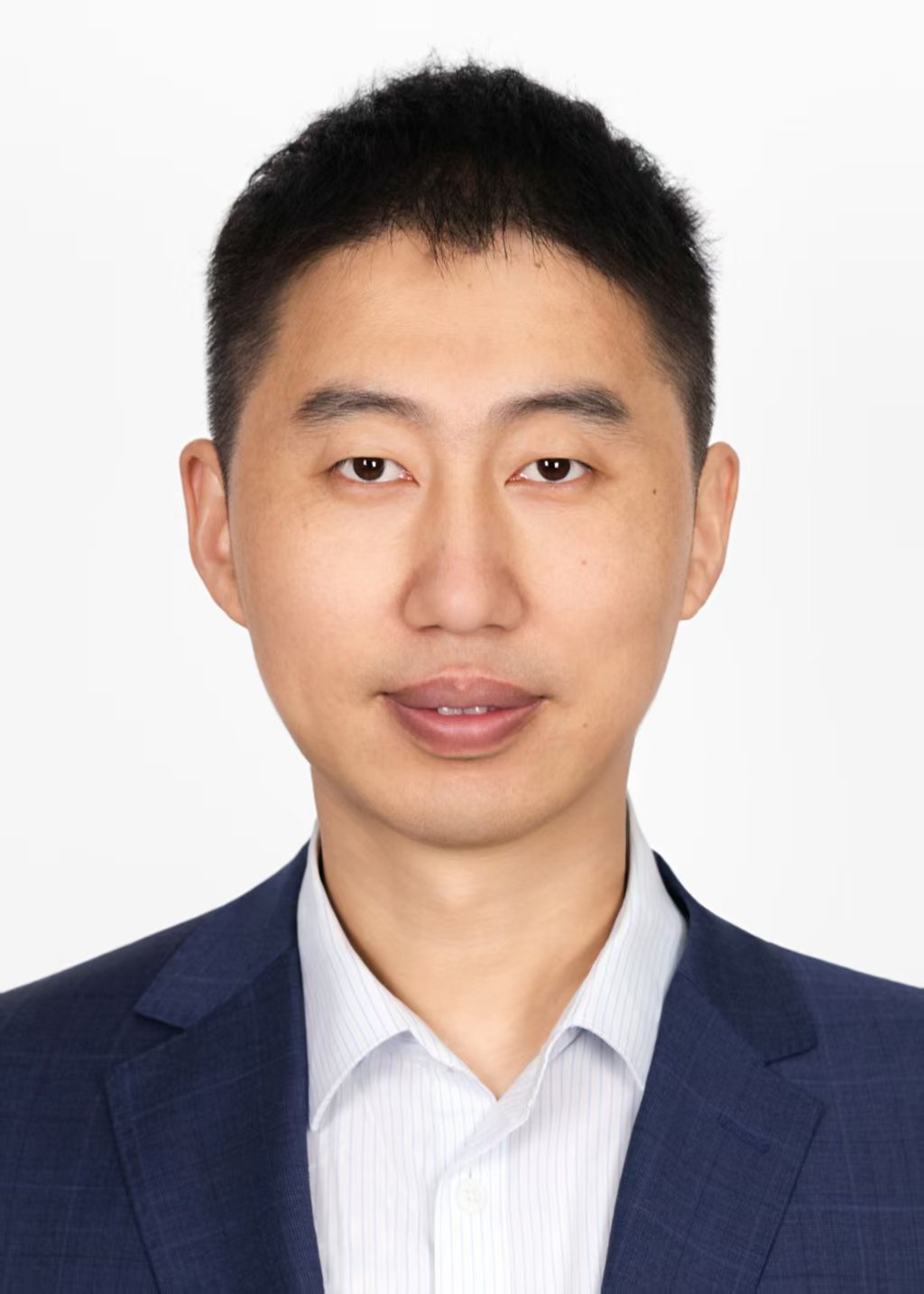}}] {Tongwei Ren} (Member, IEEE) received the B.S., M.E., and Ph.D. degrees from Nanjing University, Nanjing, China, in 2004, 2006, and 2010, respectively. He is currently a Professor with the School of Artificial Intelligence, China Pharmaceutical University, Nanjing, China. His research interest mainly includes multimedia computing and its real-world applications. He has published more than 100 papers in top-tier journals and conferences. He was a recipient of the best paper candidate awards of ICIMCS 2014, PCM 2015, ICMR 2020 and MMAsia 2020, and he was in the champion teams of ECCV 2018 PIC, MM 2019 VRU, MM 2020 DVU, MM 2022 DVU, and MM 2023 DVU challenges.
 \end{IEEEbiography}

 \begin{IEEEbiography}[{\includegraphics[trim=6.75bp 14.0625bp 0bp 2.8125bp,clip,width=1in,height=1.25in,keepaspectratio]{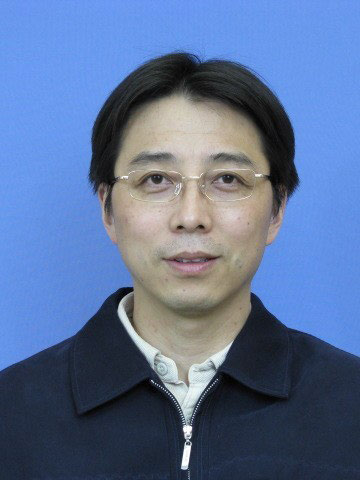}}] {Gangshan Wu} (Member, IEEE) received the B.Sc., M.S., and Ph.D. degrees from the Department of Computer Science and Technology, Nanjing University, Nanjing, China, in 1988, 1991, and 2000, respectively. He is currently a Professor with the School of Computer Science, Nanjing University. His current research interests include computer vision, multimedia content analysis, multimedia information retrieval, digital museum, and large-scale volumetric data processing.
 \end{IEEEbiography}

\vfill

\end{document}